\documentclass[runningheads]{llncs}

\usepackage{eccv}

\usepackage{eccvabbrv}

\usepackage{graphicx}
\usepackage{booktabs}

\usepackage{amsmath}
\usepackage{amssymb}
\usepackage{booktabs}
\usepackage{listings}
\usepackage{enumitem}
\usepackage{verbatim}
\usepackage{pifont}
\usepackage{amssymb}
\usepackage{dblfloatfix}
\usepackage{footmisc}
\usepackage{float}
\usepackage{dblfloatfix}
\usepackage{colortbl}
\usepackage[utf8]{inputenc}
\usepackage{amsmath}
\usepackage{algorithm}
\usepackage{algpseudocode}
\usepackage[toc,page]{appendix}

\usepackage[accsupp]{axessibility}  
\usepackage{multirow}

\usepackage{pifont}
\newcommand{\xmark}{\ding{55}}%
\newcommand{\cmark}{\ding{51}}%
\definecolor{Gray}{gray}{0.9}
\usepackage{amsmath}

\definecolor{gray9}{gray}{.9}
\definecolor{gray95}{gray}{.95}
\definecolor{gray8}{gray}{.8}
\definecolor{gray85}{gray}{.85}

\usepackage[pagebackref,breaklinks,colorlinks,citecolor=eccvblue]{hyperref}
\usepackage[capitalize]{cleveref}
\crefname{section}{Sec.}{Secs.}
\Crefname{section}{Section}{Sections}
\Crefname{table}{Table}{Tables}
\crefname{table}{Tab.}{Tabs.}

\usepackage{orcidlink}
\usepackage{refcount}

\begin{document}

\title{PosSAM: Panoptic Open-vocabulary \\ Segment Anything} 

\titlerunning{PosSAM}
\author{Vibashan VS\thanks{These authors contributed equally to this work.}\inst{1, 2}\orcidlink{0000-0003-4766-2901} \and
Shubhankar Borse$^*$\inst{1}\orcidlink{0009-0009-9171-4431} \and
Hyojin Park\inst{1}\orcidlink{0000-0001-7263-8298} \and 
Debasmit Das\inst{1}\orcidlink{0000-0003-2910-5079} \and
Vishal Patel\inst{2}\orcidlink{0000-0002-5239-692X} \and
Munawar Hayat\inst{1}\orcidlink{0000-0002-2706-5985} \and
Fatih Porikli\inst{1}\orcidlink{0000-0002-1520-4466} }

\authorrunning{V. VS, S.Borse, H. Park, D. Das, V. Patel, M. Hayat and F. Porikli}

\institute{Qualcomm AI Research\thanks{Qualcomm AI Research, an initiative of Qualcomm Technologies, Inc.} \and
Johns Hopkins University}


\maketitle

\setlength{\tabcolsep}{10pt} 

\begin{abstract}

In this paper, we introduce an open-vocabulary panoptic segmentation model that effectively unifies the strengths of the Segment Anything Model (SAM) with the vision-language CLIP model in an end-to-end framework. While SAM excels in generating spatially-aware masks, it's decoder falls short in recognizing object class information and tends to oversegment without additional guidance. Existing approaches address this limitation by using multi-stage techniques and employing separate models to generate class-aware prompts, such as bounding boxes or segmentation masks.  Our proposed method, PosSAM is an end-to-end model which leverages SAM's spatially rich features to produce instance-aware masks and harnesses CLIP's semantically discriminative features for effective instance classification. Specifically, we address the limitations of SAM and propose a novel Local Discriminative Pooling (LDP) module leveraging class-agnostic SAM and class-aware CLIP features for unbiased open-vocabulary classification. Furthermore, we introduce a Mask-Aware Selective Ensembling (MASE) algorithm that adaptively enhances the quality of generated masks and boosts the performance of open-vocabulary classification during inference for each image. We conducted extensive experiments to demonstrate our methods strong generalization properties across multiple datasets, achieving state-of-the-art performance with substantial improvements over SOTA open-vocabulary panoptic segmentation methods. In both COCO to ADE20K and ADE20K to COCO settings, PosSAM outperforms the previous state-of-the-art methods by a large margin, 2.4 PQ and 4.6 PQ, respectively. Project Website: \href{https://vibashan.github.io/possam-web/}{https://vibashan.github.io/possam-web/}.

\end{abstract}

\section{Introduction}

Segment Anything Model (SAM) \cite{kirillov2023segment} shows remarkable promptable mask generation capabilities. While the existing approaches struggle with generalization to new objects, SAM can effectively generalize to a variety of visual concepts and image distributions, thanks to its large-scale pretraining using 1 billion masks. Instead of relying on costly human annotations, SAM plays a dual role as a data engine to generate and refine its masks during training iteratively. 

\begin{figure}[!t]
    \centering
    \includegraphics[width=1.0\linewidth]{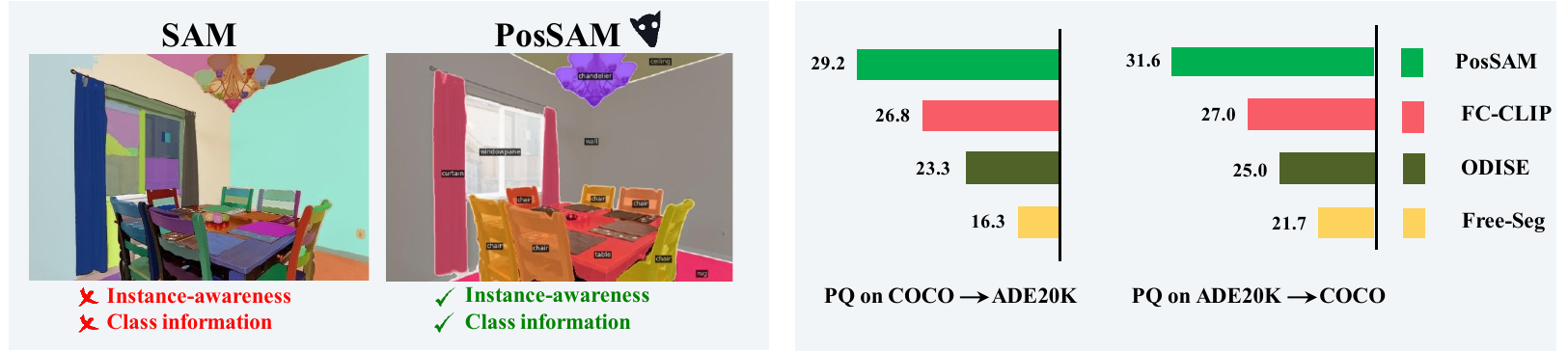}
    \caption{\textbf{Left}: While SAM possesses exceptional spatial awareness and promptable segmentation capabilities, it lacks class/semantic awareness and tends to over-segment objects into multiple regions. Our proposed PosSAM enhances SAM with instance and class awareness by efficiently integrating SAM's representations with semantically discriminative CLIP embeddings, resulting in robust open-vocabulary panoptic segmentation. \textbf{Right}: As shown, we achieve state-of-the-art performance in both COCO to ADE20K and ADE20K to COCO settings. 
    }
    \label{fig:intro}
\vspace{-2mm}
\end{figure}

Despite its potential, SAM faces a significant limitation in generating instance and class-aware masks, which are vital for various downstream segmentation tasks.  For example, as illustrated in Fig.~\ref{fig:intro}, SAM tends to over-segment objects, resulting in masks that lack instance-awareness. Further, SAM~\cite{kirillov2023segment} has proposed text prompting to predict class-aware masks; however, the paper stated that the current SAM can work only with simple prompts targeting a single class, thereby limiting its applicability to only semantic segmentation. To address these instance and class-aware limitations, existing approaches resort to specialized networks to obtain more capable prompts. For example, GroundingDINO \cite{liu2023grounding} initially predicts bounding boxes, while Semantic Segment-Anything (SSA) \cite{chen2023semantic} concurrently employs a segmentation model. Similarly, the recently introduced Semantic-SAM \cite{li2023semantic} relies on a segmentation model \ie Mask-DINO \cite{li2023mask}. Although these methods offer potential, their additional disjoint steps lead to inefficiency and a lack of end-to-end awareness. Moreover, the bounding box-based methods can only generate instance masks and lack the ability to perform panoptic segmentation. Our method addresses these limitations by introducing a unified end-to-end trainable approach that leverages SAM's rich spatial features for generating class-agnostic panoptic masks in an open-vocabulary setting, enabling SAM for downstream segmentation tasks.

\begin{figure*}[t]
  \centering
  \includegraphics[width=1.0\linewidth]{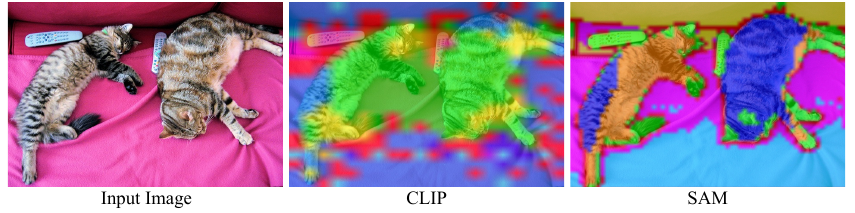}
   \caption{Visualization of K-means clustering of frozen CLIP \cite{radford2021learning} and SAM \cite{kirillov2023segment} backbone features. As illustrated, SAM's clustering maps show a higher precision in object localization when compared to the cluster map of CLIP. While SAM lacks instance awareness, it still produces more defined boundaries between parts of objects, indicating its enhanced spatial awareness and fine-grained representation learning capabilities.}
   \label{fig:motive}
\vspace{-2.5mm}
\end{figure*}

Most open-vocabulary segmentation methods~\cite{zhou2023zegclip,li2024omg,ma2022open,zou2023generalized,xu2023side,zhou2022extract,ding2022decoupling,liang2023open} rely on CLIP \cite{radford2021learning} for alignment, mask generation or classification. For instance, MaskCLIP \cite{ding2022open} and OVSeg \cite{liang2023open} adopt a two-stage approach by adapting frozen CLIP for alignment and classification. Similarly, FC-CLIP \cite{yu2023convolutions} and OMG-Seg \cite{li2024omg} leverage a frozen CLIP backbone for mask-generation as well as classification. However, we argue that leveraging SAM for mask generation due to its rich feature representation.
To illustrate this, we compare feature clustering results from different backbones in Fig.~\ref{fig:motive}. SAM's cluster map demonstrates more precise object boundaries compared to CLIP. This motivates us to leverage SAM as a backbone for class-agnostic mask generation due to its strong spatial and localization awareness. Additionally, we empirically observed that the SAM decoder generates poor panoptic masks (see Table.\ref{tab:decoder}). Therefore, following prior OV works \cite{yu2023convolutions,li2024omg,xu2023odise,zou2023generalized,liang2023open}, we employ Mask2former as our mask decoder.


PosSAM (pronounced \textit{possum}) addresses the key challenges to enable SAM for open-vocabulary segmentation by generating instance-aware masks, classifying class-agnostic SAM features, and effectively distinguishing between seen (in-vocabulary) and unseen (out-of-vocabulary) categories. We leverage a feature pyramid network to project SAM ViT features into hierarchical multi-scale feature maps, which are then used by the mask decoder to obtain instance-aware masks. As our backbone remains fixed during training, we introduce an auxiliary IoU prediction task to regularize generated masks for instance-awareness. For OV classification, solely relying on discriminative CLIP features makes the model biased towards the training set i.e. seen classes. To mitigate this bias, we introduce a Local Discriminative Pooling (LDP) module (Sec.~\ref{sec:maskclass}) that facilitates the interaction between discriminative CLIP features and class-agnostic SAM features, thereby avoiding overfitting to seen classes during training. During inference, we propose a novel Mask-Aware Selective Ensemble (MASE) algorithm which leverages both IoU scores and LDP confidence scores to adaptively distinguish between in-vocabulary and out-of-vocabulary classes for robust real-world OV segmentation. Our proposed method demonstrates state-of-the-art performance across various open-vocabulary segmentation tasks.

Our contributions can be summarized as:
\begin{itemize}
    \item We introduce PosSAM, an open-vocabulary panoptic segmentation model that generates class and instance-aware masks with excellent generalization to a variety of visual concepts by unifying SAM and CLIP in an end-to-end trainable framework.
    \item We develop a novel Local Discriminative Pooling (LDP) module to enhance discriminative CLIP features with class-agnostic SAM features for an unbiased OV classification.
    \item We introduce the Mask-Aware Selective Ensembling algorithm to adaptively discern between seen and unseen classes by leveraging IoU and LDP confidence scores for each image.
    \item We conduct extensive experiments and demonstrate superior performance over existing state-of-the-art open-vocabulary panoptic segmentation methods across multiple benchmark datasets. 
\end{itemize}

\section{Related Work}
\label{RelatedWork}

\subsection{Vision-Language Models}
Vision-Language Models (VLM) have been quite effective in learning a unified representation space for holistic understanding of both modalities~\cite{radford2021learning, furst2022cloob, singh2022flava}. These models normally consist of an image encoder, a text encoder and fusion module to combine information from each encoder. The most common approach for training these models is contrastive learning which uses large-scale datasets of image and caption pairs to minimize a distance of embedded features between positive pairs and maximize between negative pairs~\cite{radford2021learning,vs2023towards,furst2022cloob,jia2021scaling,vs2023instance,li2021supervision,zhai2022lit,singh2022flava}.  
Another category of models uses prefix by predicting next text sequence from given image patch sequence and prefix text sequence ~\cite{wang2021simvlm,tsimpoukelli2021multimodal,alayrac2022flamingo,manas2022mapl}.
Similarly, \cite{xu2023bridgetower, tan2019lxmert, lu2019vilbert, li2019visualbert} predict masked words based on the corresponding image or predict whether given text and images are aligned or not.

\subsection{Open-vocabulary segmentation}

Deep learning has been successfully applied for various dense prediction tasks such as image segmentation \cite{long2015fully}, with hierarchical branches \cite{deeplabV2,deeplabV3,zhao2017pspnet}, attention mechanisms \cite{yuan2018ocnet,huang2019ccnet,borse2023x, borse2022panoptic,strudel2021segmenter}, and auxiliary losses \cite{borse2021hs3,borse2021inverseform,borse2023dejavu}, to point out a few. Open-vocabulary segmentation is a recent task in image segmentation to gauge generalization ability to novel visual categories that have not been encountered during training \cite{li2022languagedriven,ma2022open,zhou2023zegclip,xu2023side,xu2022groupvit,vs2023mask,ding2022decoupling,xu2021simple,ghiasi2022scaling}. 
Most works~\cite{li2024omg,xu2023masqclip,chen2023open,xu2023open,qin2023freeseg,ding2022open} in open-vocabulary segmentation use features from large pre-trained multi-modal foundational models such as CLIP~\cite{radford2021learning} and leverage Mask2former \cite{cheng2022masked} as decoder. MaskCLIP~\cite{ding2022decoupling} recently builds upon the CLIP model~\cite{radford2021learning} to open-vocabulary panoptic segmentation. This is achieved through a two-stage procedure consisting of a mask generator and CLIP encoder for alignment between the two modalities. ODISE~\cite{xu2023open} uses the representation space of text-image diffusion models to generalize to novel classes for open-vocabulary segmentation. OPSNet~\cite{chen2023open} proposes an embedding modulation by merging mask query and CLIP information by linear fusion using a similarity score obtained between new categories and known training categories. FreeSeg~\cite{qin2023freeseg} enforces integration of multiple granular concepts into a textual description, enabling generalizability to novel categories. In an alternative work, FCCLIP~\cite{yu2023convolutions} proposes a single-stage framework by using a frozen convolution-based CLIP backbone. While existing SOTA approaches for open vocabulary panoptic segmentation employ the features of contrastively trained CLIP \cite{radford2021learning} or generative diffusion models. However, the spatially rich fine-grained feature space of the SAM model for open-vocabulary panoptic segmentation has not yet been explored. Our paper is an effort in this direction and learns to equip SAM with instance and class awareness and discussed next.

\subsection{Segment Anything}
Multiple works extend SAM~\cite{kirillov2023segment} to different use cases. These include applying on medical images~\cite{ma2023segment,zhang2023input}, camouflaged objects~\cite{tang2023can} and transparent objects~\cite{han2023segment}, and make it class aware. A recent approach extends SAM~\cite{groundedsam}, by combining Grounding DINO~\cite{liu2023grounding} with SAM, which allows detection and segmentation of objects using text input.~\cite{zhang2023faster} develops a faster version of SAM for edge devices by replacing the heavyweight Vision Transformer (ViT)~\cite{dosovitskiy2020image} with lightweight ViT~\cite{wu2022tinyvit}. Since SAM does not generate class-label predictions, other studies~\cite{park2023segclip, chen2023semantic} integrate SAM with models like BLIP~\cite{li2022blip} or CLIP~\cite{radford2021learning} and improve segmentation accuracy. Furthermore, SAM has been used in works beyond object segmentation such as image editing~\cite{rombach2022high}, inpainting~\cite{yu2023inpaint}, video object tracking~\cite{yang2023track,cheng2023segment}, etc. SAM has also been used to facilitate semantic communication~\cite{tariq2023segment} by transmitting only foreground objects. It has also been used for 3D reconstruction~\cite{shen2023anything,kang2022any} from single images. Extending SAM for the task of OV panoptic segmentation still remains a challenging problem to solve. 

\section{Method}
\label{sec:method}

\subsection{Problem Setting}
\label{sec:problem}
Consider an input image $\mathbf{I}\in\mathbb{R}^{\mathcal{H\times W}\times3}$ with it's corresponding binary mask labels $\mathbf{Y_{mask}} \in \mathcal{M}^{\mathcal{H\times W}\times N_{mask}}$ and class labels $\mathbf{Y_{class}} \in \mathbf{\mathcal{C}}_{train}$. Here, $\mathcal{M}=\left\lbrace0, 1\right\rbrace$ is the set of mask labels, $N_{mask}$ is the number of ground truth masks corresponding to the image $\mathbf{I}$ and $\mathbf{\mathcal{C}}_{train}$  is set of training categories. Our objective is to train a network utilizing the images and labels in the training data, and consequently perform panoptic segmentation on an unseen test image $\mathbf{J}\in\mathbb{R}^{\mathcal{H\times W}\times3}$ which may contain objects from both seen and unseen classes \ie $\mathbf{\mathcal{C}}_{test}$. 

\subsection{Panoptic Open-vocabulary Segment Anything (PosSAM)}
\label{sec:overview} 
\begin{figure*}[t]
  \centering
  \includegraphics[width=1.0\linewidth]{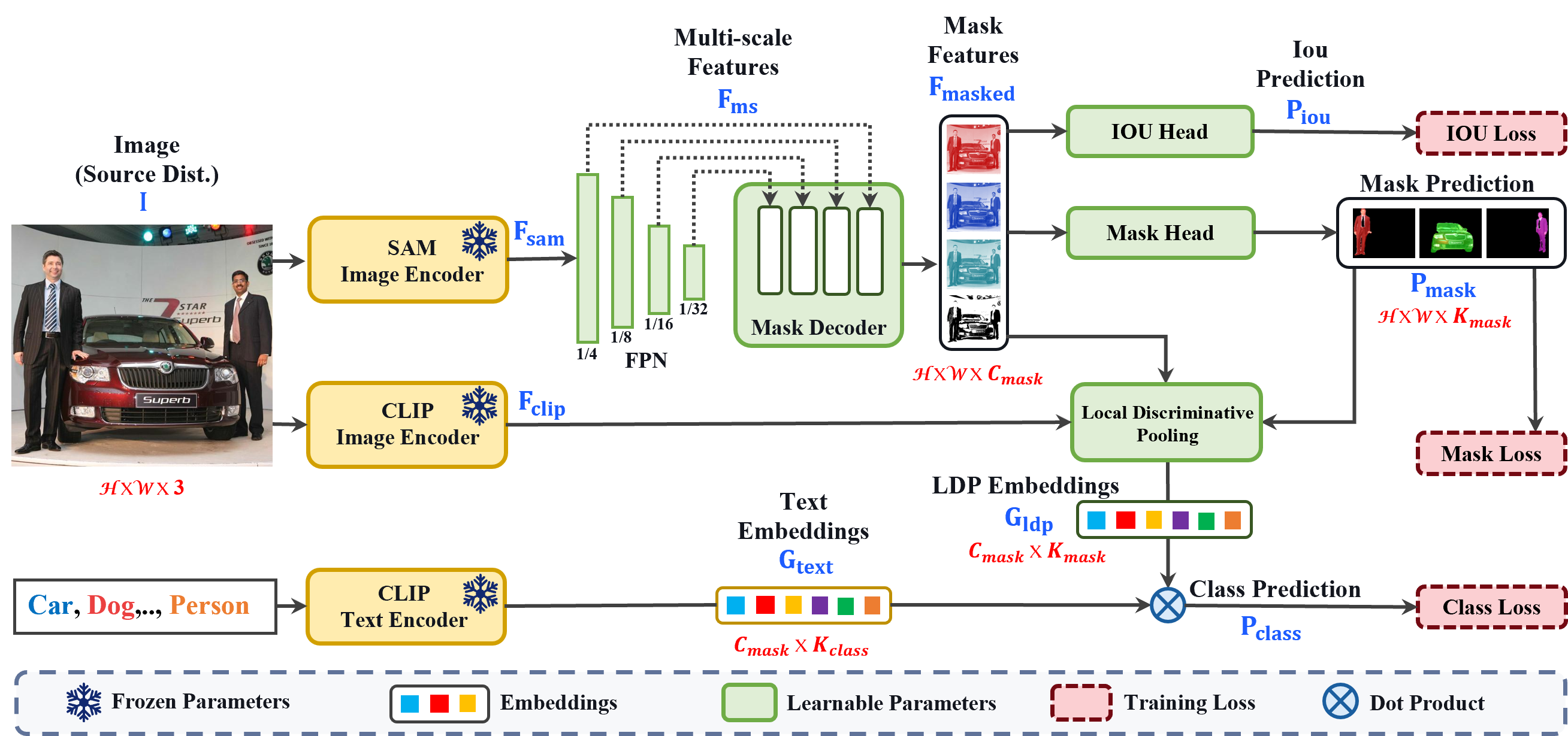}
   \caption{Overview of our PosSAM training pipeline. We first encode the input image using the SAM backbone to extract spatially rich features, which are processed through a Feature Pyramid Network to obtain hierarchical multi-scale features decoded to form mask features and predict class-agnostic masks. Concurrently, we train an IoU predictor for each mask to measure its quality. We obtain CLIP image features and develop our proposed LDP module to achieve better classification of these masks. We learn to pool and generate enhanced class-specific features, which are then classified by a process supervised with ground truth category labels derived from the CLIP text encoder.}
   \label{fig:training}
\end{figure*}

\begin{figure*}[t]
  \centering
  \includegraphics[width=1.0\linewidth]{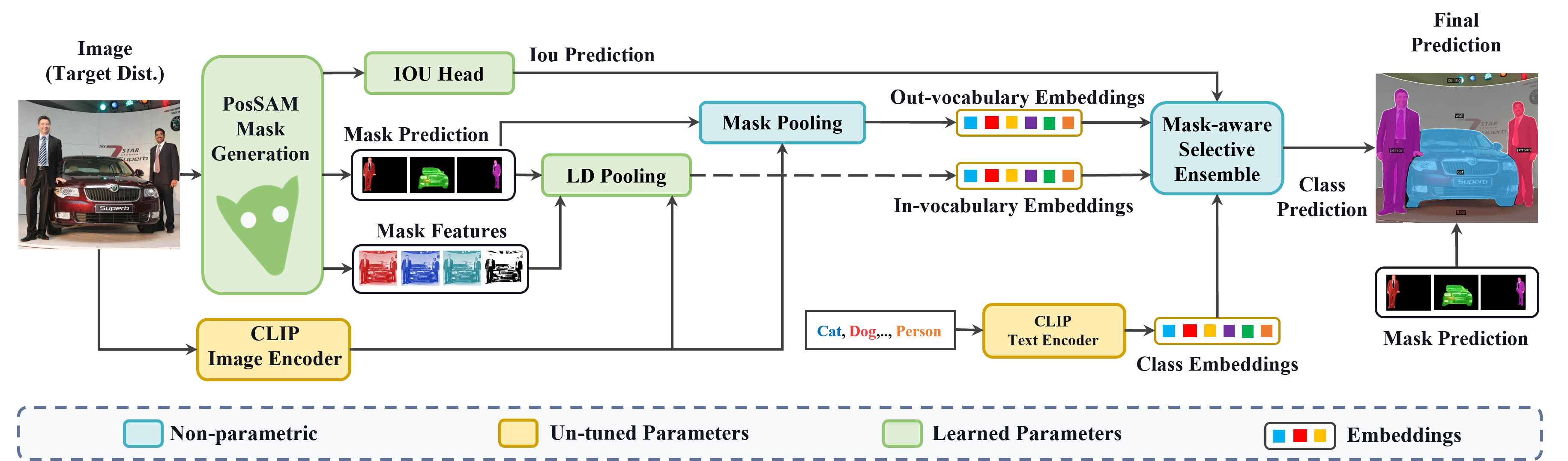}
   \caption{ In the inference pipeline, the LDP embeddings and CLIP embeddings are generated from the local discriminative pooling module and mask pool module, respectively. These embeddings are used to classify the mask proposal by performing a product with pre-computed CLIP text embeddings.  Final predictions are processed by our MASE strategy, where IoU score is utilized to weigh the classification predictions and an adaptive geometric ensemble is applied to the outputs of the LDP and CLIP embeddings.}
   \label{fig:inference}
\end{figure*}

We illustrate our proposed training pipeline in Fig.~\ref{fig:training} and our inference pipeline in Fig.~\ref{fig:inference}. As shown in Fig.~\ref{fig:training}, we first pass the input image through the SAM~\cite{kirillov2023segment} image encoder to generate features $\mathbf{{F}_{sam}}$. We keep the SAM encoder frozen during training.
Following this, $\mathbf{{F}_{sam}}$ are passed through a multi-scale FPN block described in Sec.~\ref{sec:maskgen}, generating hierarchical features $\mathbf{{F}_{ms}}$, which are then fed to a mask decoder~\cite{cheng2021masked} which generates intermediate mask features $\mathbf{{F}_{masked}}$, used by mask head to get predicted masks $\mathbf{P_{mask}} \in \mathcal{M}^{\mathcal{H\times W}\times K_{mask}}$. Here, $K_{mask}$ is the total number of predicted masks. The mask features are also fed into an IoU head to predict the IoU $\mathbf{P_{iou}}$ corresponding to each mask. A detailed explanation of the IoU head is in Sec.~\ref{sec:maskgen}. 

To generate class-aware image features $\mathbf{{F}_{clip}}$, we pass the image $\mathbf{I}$ through the CLIP~\cite{gupta2020contrastive} image encoder, which is kept frozen during training. The generated clip features are then fed along with mask features $\mathbf{{F}_{masked}}$ and predicted masks $\mathbf{P_{mask}}$ to our proposed Local Discriminative Pooling block. This block is illustrated in Fig.~\ref{fig:ldp_block} and described in Sec.~\ref{sec:maskclass} and produces the in-vocabulary pooled embeddings $\mathbf{G_{ldp}}$, illustrated in both Fig.~\ref{fig:training} and Fig.~\ref{fig:inference}. In the following, we describe our mask generation pipeline in detail. Once we generate the masks and in-vocabulary pooled embeddings, we follow separate protocols for training and inference, discussed in Sec.~\ref{sec:maskclass} and Sec.~\ref{sec:inference} respectively.

\subsection{Mask Generator}
\label{sec:maskgen}

\noindent \textbf{FPN:} SAM's ViT produces single-resolution feature maps, which are sub-optimal for segmenting objects of varying sizes. To learn a hierarchical representation \cite{li2022exploring},  we design an FPN network, with a combination of convolutions and deconvolution layers, and feed $\mathbf{{F}_{sam}}$ through it to generate feature maps  $\mathbf{{F}_{ms}}$ at different scales of $\frac{1}{32}$, $\frac{1}{16}$, $\frac{1}{8}$, and $\frac{1}{4}$. These hierarchical features preserve ViT's robustness while providing the necessary multi-scale representation.

\noindent \textbf{Mask Decoder:} Following Mask2Former \cite{cheng2021masked}, the mask decoder takes the  multi-scale feature maps $\mathbf{{F}_{ms}}$ and enhances it using the pixel decoder. These enhanced pixel features, along with object queries \cite{carion2020end}, undergo a sequence of mask decoding steps, including masked cross-attention, self-attention, and an FFN and output region features $\mathbf{F_{masked}}$, which are then fed into the mask head to generate class-agnostic binary masks $\mathbf{P_{mask}}$. To train mask decoder, we use Hungarian matching \cite{carion2020end} to establish one-to-one correspondences between predicted and ground-truth masks and we supervise the predicted class-agnostic binary masks via a pixel-wise binary cross entropy loss as follows:
\begin{equation}
    \mathcal{L}_{mask}(i,j) =  BCE(\mathbf{Y_{mask}^{i}}, \mathbf{P_{mask}^{j}})
\end{equation}
where, $i$ and $j$ are Hungarian \cite{carion2020end} matched indices obtained by matching ground-truth mask $\mathbf{Y_{mask}} \in \mathcal{M}^{\mathcal{H\times W}\times N_{mask}}$  and predicted masks $\mathbf{P_{mask}} \in \mathcal{M}^{\mathcal{H\times W}\times K_{mask}}$.

\noindent \textbf{Intersection over Union (IoU) Head:} In addition to the mask generation process, we integrate an IoU head into our mask decoder. The IoU head takes output region features  $\mathbf{F_{masked}}$ as input and estimates the IoU score between predicted binary masks and their corresponding ground truths. The IoU head is trained for Hungarian-matched predicted and ground-truth masks as follows:
\begin{equation}
    \mathcal{L}_{iou}(i,j) =  \| \mathbf{Y_{mask}^{i}} - \mathbf{P_{mask}^{j}}\|_2^2
\end{equation}

The IoU head is trained in a category-agnostic manner and can easily generalize to unseen classes to gauge the quality of the generated masks. Therefore during inference, the predicted IoU scores are utilized to weigh the logit scores, giving more importance to the confident masks.

\vspace{-2mm}
\subsection{Mask Classification}
\label{sec:maskclass}

We leverage CLIP~\cite{radford2021learning} text encoder $\mathbf{c_{text}}$ to classify each of the predicted binary masks. For this, we extract class embedding corresponding to each mask by performing ROI Pool on output region features \(\mathbf{F}_{\text{masked}}\). However, since SAM features are inherently class-agnostic and the mask generator is also trained in a class-agnostic manner, we employ the CLIP image encoder to induce semantic richness. We introduce a novel learnable pooling module termed `Local Discriminative Pooling' to accumulate benefits from both CLIP and SAM representation spaces for unbiased OV classification.

\begin{figure}[!t]
    \centering
    \includegraphics[width=1.0\linewidth]{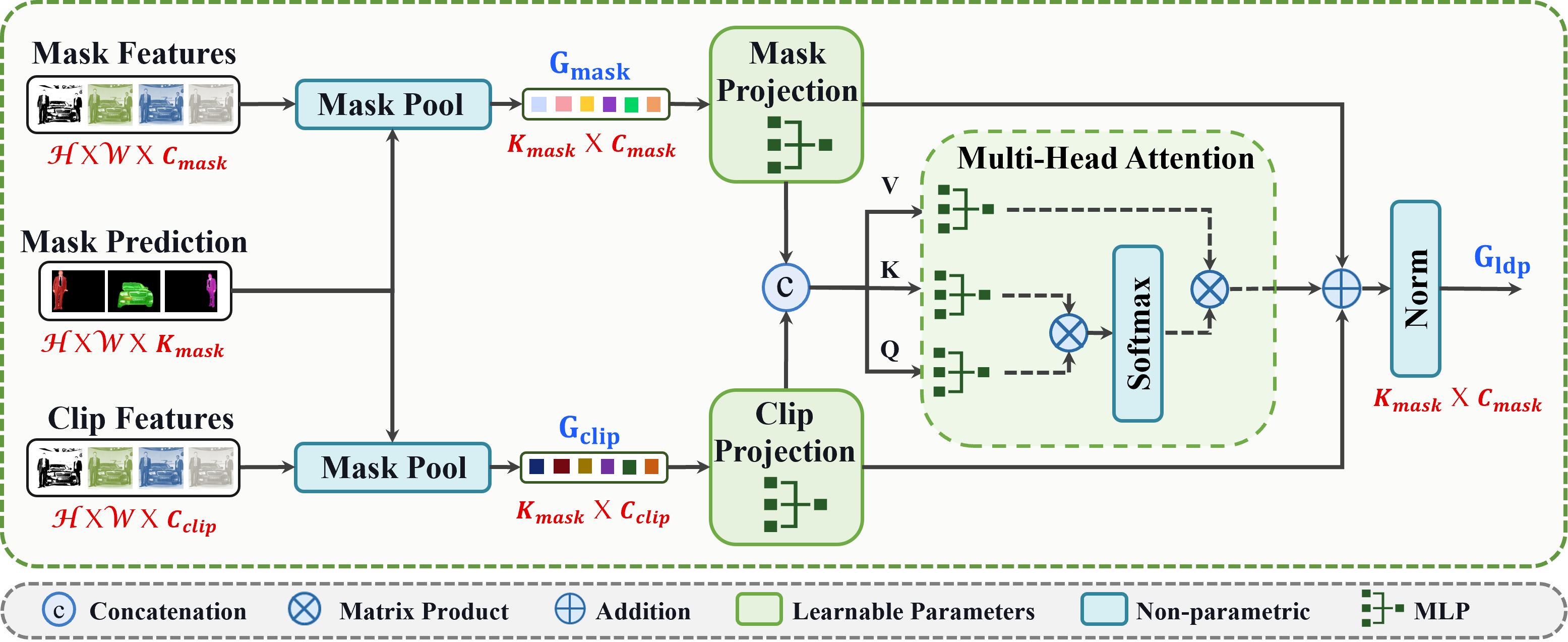}
    \caption{Local Discriminative Pooling Module. The LDP module learns to effectively fuse the class-agnostic SAM features with discriminative CLIP features, thereby avoiding overfitting to seen classes during training, which is crucial for OV inference.}
    \label{fig:ldp_block}
    \vspace{- 1.2 em}
\end{figure}

\noindent{\bf{Local Discriminative Pooling (LDP):}} As illustrated in Fig.~\ref{fig:ldp_block}, we introduce a dual-stream architecture that concurrently processes mask and clip features. We first apply ROI Pooling (i.e. mask pooling) on both the SAM and CLIP features and then project them to align with the CLIP \cite{radford2021learning} and SAM \cite{kirillov2023segment} embedding space. These features are then concatenated and fused using self-attention. Finally, the refined features are passed through the normalization layer to obtain the locally discriminative pooled embeddings $\mathbf{G_{ldp}}$, (LDP embeddings) for each mask. The LDP module generates both semantically rich (attributed to precise SAM features) as well as class discriminative features (attributed to CLIP).

\noindent \textbf{Classification loss:} 
For each category present in $\mathcal{C}_{train}$, we obtain the corresponding CLIP text embedding $\mathbf{G_{text}}$. Next, we compute the probability of LD pooled embedding as $\mathbf{P_{class}} = \text{Softmax}(\mathbf{G_{ldp}} \cdot \mathbf{G_{text}}/\tau)$, which is used to compute cross-entropy loss: 

\begin{equation}
\setlength{\belowdisplayskip}{1pt}
\setlength{\abovedisplayskip}{1pt} \setlength{\abovedisplayshortskip}{1pt}
\mathcal{L}_{class} (i,j) = \text{CE}(\mathbf{Y_{class}}^i, \mathbf{P_{class}}^j),
\end{equation}

where $\tau$ is a learnable temperature parameter and $\mathbf{P_{class}}$ are the predicted class logits. The final loss used for training is a weighted combination the three proposed loss functions,
\begin{equation}
\setlength{\belowdisplayskip}{1pt} \setlength{\belowdisplayshortskip}{1pt}
\setlength{\abovedisplayshortskip}{1pt}
\mathcal{L}_{total} (i,j) = \gamma_{a} \mathcal{L}_{class} (i,j) + \gamma_{b} \mathcal{L}_{mask} (i,j) + \gamma_{c} \mathcal{L}_{iou} (i,j),
\end{equation}

where the $\gamma$ values are loss weight and are considered hyperparameters. In the following section, we provide a detailed description of our inference pipeline. 

\subsection{Open-Vocabulary Inference}
\label{sec:inference}
As illustrated in Fig.~\ref{fig:inference}, we adapt our pipeline during inference to accommodate for the unseen classes. Since our model produces accurate masks, we fix the mask generation pipeline. However, we adapt the classification branch to better generalize to previously unseen classes.

\noindent{\bf{LDP and CLIP Classification:}} As our LDP module is trained on a closed set of training categories $\mathcal{C}_{train}$, we use the same module to generate pooled embeddings corresponding to categories in the set $\mathcal{C}_{train} \cap \mathcal{C}_{test}$ which contains object classes the model has seen before. These embeddings are denoted as LDP embeddings, $\mathbf{G_{ldp}}$. To allow for precise prediction of the categories which the model has never seen before \ie those which fall outside the set $\mathcal{C}_{test} \cap \mathcal{C}_{train}$, we apply the ROI pooling (mask pooling) operation on the frozen CLIP to generate the CLIP embeddings $\mathbf{G_{clip}}$. Both these embeddings along with the CLIP text embeddings are fed into our proposed Mask-Aware Selective Ensemble (MASE) strategy to obtain the final predicted class logits.

\begin{algorithm}[ht]
\caption{Mask-Aware Selective Ensemble (MASE) for PosSAM}
\begin{algorithmic}[1]
\State $\mathbf{P_{iou}}$: Predicted IoU scores
\State $\mathbf{P_{class, ldp}}$: LDP logits
\State $\mathbf{P_{class, clip}}$: CLIP logits
\State $\mathcal{M}$: Category Overlapping Mask
\State $\alpha, \beta$: Ensemble weighting parameters
\Function{MASE}{}
\State $\mathbf{P_{ldp}} \leftarrow \mathbf{P_{class, ldp}} \cdot \mathbf{P_{iou}}$
\State $\mathbf{P_{clip}} \leftarrow \mathbf{P_{class, clip}} \cdot \mathbf{P_{iou}}$
\State $\tilde{\mathbf{SE}}_{\mathbf{ldp}} \leftarrow 1 - \left(\frac{\text{sort}(\mathbf{P_{ldp}})[-2]}{\text{sort}(\mathbf{P_{ldp}})[-1]}\right)$
\State $\tilde{\mathbf{SE}}_{\mathbf{clip}} \leftarrow 1 - \left(\frac{\text{sort}(\mathbf{P_{clip}})[-2]}{\text{sort}(\mathbf{P_{clip}})[-1]}\right)$
\State $\mathbf{SE}_{\mathbf{conf}} \leftarrow \frac{\tilde{\mathbf{SE}}_{\mathbf{clip}}}{\tilde{\mathbf{SE}}_{\mathbf{clip}} + \tilde{\mathbf{SE}}_{\mathbf{ldp}}}$
\State $\hat{\alpha} \leftarrow \alpha \cdot (1 + \mathbf{SE}_{\mathbf{conf}})$
\State $\hat{\beta} \leftarrow  \beta \cdot (1 + \mathbf{SE}_{\mathbf{conf}})$

\State Initialize $\mathbf{P}_{\mathbf{class}}$ to an empty structure
    \If{IoU score of all masks is valid}
        \State $\mathbf{P}_{\mathbf{in-voc}} \leftarrow (\mathbf{P_{ldp}}^{(1-\hat{\alpha})} \cdot \mathbf{P}_{\mathbf{clip}}^{\hat{\alpha}}) \odot \mathcal{M}$
        \State $\mathbf{P}_{\mathbf{out-voc}} \leftarrow (\mathbf{P_{ldp}}^{(1-\hat{\beta})} \cdot \mathbf{P}_{\mathbf{clip}}^{\hat{\beta}}) \odot (1-\mathcal{M})$
        \State  $\mathbf{P}_{\mathbf{class}} \leftarrow \mathbf{P}_{\mathbf{in-voc}} + \mathbf{P}_{\mathbf{out-voc}}$
    \EndIf
\State \Return $\mathbf{P}_{\mathbf{class}}$
\EndFunction
\end{algorithmic}
\label{algo:mase}
\end{algorithm}

\noindent{\bf{Mask-aware Selective Ensemble (MASE):}} As depicted in Algorithm~\ref{algo:mase}, MASE is a regularized and weighted aggregation strategy to generate class logits by differentiating between seen and unseen classes, effectively leveraging IoU scores, along with the LDP and CLIP embeddings. Both LDP and CLIP embeddings denoted as $\mathbf{G_{ldp}}$ and $\mathbf{G_{clip}}$ respectively, contain mask encodings which represent a vector for each mask. These are aggregated with the text embeddings $\mathbf{G_{text}}$ to generate LDP logits $\mathbf{P_{class, ldp}}$ and CLIP logits $\mathbf{P_{class, clip}}$ respectively, as described in Sec.~\ref{sec:maskclass}. Essentially, these class logits denote a distribution for each mask. Since the number of masks generated by our decoder is much higher than the number of things and stuff in an image, multiple generated masks are void. Analogous to this observation, our predicted IoU scores is a measure of the presence of valid masks. Hence, we use these IoU scores $\mathbf{P_{iou}}$ as a regularizer to weigh our predicted class logits along the mask dimension.

To combine the LDP and CLIP class predictions, we apply a geometric mean following~\cite{xu2023odise} and~\cite{yu2023convolutions} with $\alpha$ and $\beta$ as geometric mean weighting coefficients. Different from prior works, we do not fix the weights for the geometric mean and vary them class-wise based on the confidence of each prediction \ie LDP prediction receives a higher weight for a particular mask if the confidence of LDP class prediction is higher than CLIP class prediction and vice-versa for CLIP class predictions. Therefore, $\alpha$ and $\beta$ are adaptively adjusted based on the relative confidence between the LDP and CLIP classification scores to obtain the updated $\hat{\alpha}$ and $\hat{\beta}$. Once we apply the confidence-weighted and IoU-regularized geometric mean over both predictions, we attain the final class prediction $\mathbf{P_{class}}$ corresponding to each mask $\mathbf{P_{mask}}$.


\section{Experiments}
\label{sec:experiments}

\subsection{Implementation Details}
\label{sec:implementation}
\noindent \textbf{Architecture.}\quad
We use pre-trained SAM~\cite{kirillov2023segment} ViT as our backbone and project its features to a hierarchical feature map using our proposed FPN. B,L,H are the Base, Large, and Huge variants of SAM, respectively. We choose Mask2Former~\cite{cheng2021masked} as our mask decoder, to generate \(N=250\) class-agnostic mask predictions. We use CLIP~\cite{cherti2023reproducible} as our image-text discriminative model. 

\begin{figure*}[t]
    \centering
    \includegraphics[width=1.0\linewidth]{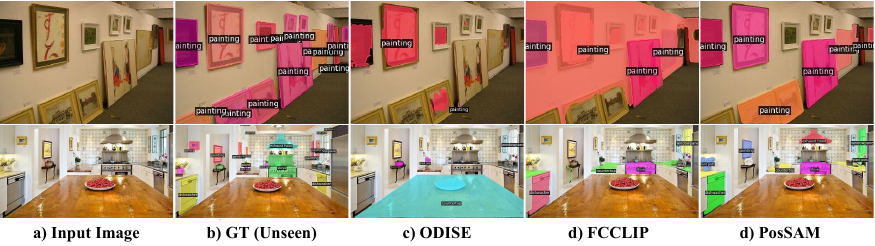}
    \caption{\textbf{Zero-shot panoptic segmentation capability from COCO to ADE20K on unseen classes}. This figure shows comparison with recent SOTA approaches. \textbf{Only novel classes are shown.} We can observe that PosSAM has the ability to accurately segment objects that are never seen before such as paintings, dishwashers, exhaust, showing advantages over ODISE \cite{xu2023odise} and FCCLIP \cite{yu2023convolutions}.}
    \label{fig:visual_novel}
    \vspace{- 1 em}
\end{figure*}

\noindent \textbf{Training Strategy.}\quad
We trained PosSAM for 250K iterations with image crops of size \( 1024 \times 1024 \), resizing the shorter side to \( 1024 \) \cite{xu2023odise,yu2023convolutions}. We trained our using 8 NVIDIA A100 GPUs, processing 2 images per GPU, for an effective batch size of 16. We use AdamW optimizer \cite{kingma2014adam, loshchilov2017decoupled}, with a weight decay of 0.05 with initial learning rate is set to $1 \times 10^{-4}$, followed by a multi-step decay schedule. The model is trained with a batch size of $16$ for a total of $50$ epochs on the COCO panoptic training dataset \cite{lin2014microsoft}.

\noindent \textbf{Inference Strategy.}\quad
We evaluate PosSAM on the ADE20K dataset~\cite{zhou2017scene} for open-vocabulary panoptic, and semantic segmentation, as well as on the Pascal datasets~\cite{everingham2010pascal, mottaghi2014role} for semantic segmentation. For panoptic segmentation, the metrics used are panoptic quality (PQ), segmentation quality (SQ), recognition quality (RQ)~\cite{kirillov2018panoptic}, Average Precision (AP), and mean intersection-over-union (mIoU), while for semantic segmentation, we use mIoU~\cite{everingham2010pascal}. In addition, for some experiments, we report on \( PQ^{\text{stuff}} \) and \( PQ^{\text{thing}} \)~\cite{kirillov2018panoptic}, differentiating between countable objects (eg: cats, dogs) and uncountable regions (eg: road, sky). All reported results are evaluated using a single checkpoint trained exclusively on the COCO panoptic training set~\cite{lin2014microsoft}. Furthermore, we implement the mask-aware selective ensembling of LDP and CLIP classification logits. Consistent with previous studies, our experiments also leverage prompt engineering techniques from~\cite{ghiasi2022scaling, xu2023open} and prompt templates from~\cite{gu2021open, liang2023open}, unless otherwise stated. See Appendix A for additional implementation details along with hyperparameters, and Appendix B for an analysis of our compute requirements.

\vspace{- 1 em}
\begin{table*}[!ht]
\label{tab:pan_ade20k}
\centering
\resizebox{1.0\linewidth}{!}{\begin{tabular}{l|ccccc|ccccc}
\toprule
& \multicolumn{5}{c|}{COCO $\rightarrow$ ADE20K} & \multicolumn{5}{c}{ADE20K $\rightarrow$ COCO} \\
Method & PQ    & SQ     & RQ     & mAP     & mIoU & PQ    & SQ     & RQ     & mAP     & mIoU \\ \toprule
MaskCLIP~\cite{ding2022open} (ICML 2023) & 15.1 & 23.7 & - & 6.0 & 23.7 & - & - & - & - & - \\
FreeSeg~\cite{qin2023freeseg} (CVPR 2023) & 16.3 & 71.8 & 21.6 & 6.5 & 24.6 & 21.7 & 72.0 & 21.6 & 6.6 & 21.7 \\
ODISE~\cite{xu2023open} (CVPR 2023)  & 23.2 & 74.4 & 27.9 & 14.0 & 28.4 & 25.0 & 79.4 & 30.4 & - & - \\
OPSNet~\cite{chen2023open} (ICCV 2023) & 17.7 & 54.9 & 21.6 & - & - & - & - & - & - & - \\
MaskQCLIP~\cite{xu2023masqclip} (ICCV 2023)   & 23.3 & - & - & - & 30.4 & - & - & - & - & - \\ 
FCCLIP~\cite{yu2023convolutions}  (NeurIPS 2023)  & 26.8 & 71.5 & 32.3 & 16.8 & 34.1 & 27.0 & 78.0 & 32.9 & - & - \\
\midrule 
\rowcolor{Gray}  PosSAM (B) &  26.5 &  74.7 &  32.0 &  16.8 &  32.9 &  25.1 &  80.1 &  30.4 &  18.7 &  37.9 \\
\rowcolor{Gray} PosSAM (L) &  28.0 &  75.5 &  33.4 &  18.3 &  34.2 &  30.7 &  \textbf{81.7} &  35.9 &  22.5 &  \textbf{41.2}  \\
\rowcolor{Gray} PosSAM (H)  &  \textbf{29.2} &  \textbf{76.3} &  \textbf{34.7} &  \textbf{18.3} &  \textbf{35.1} &  \textbf{31.6} &  81.5 &  \textbf{37.5} &  \textbf{25.9} &  39.9 \\ 

\bottomrule
\end{tabular}}
\caption{
    Zero-shot open-vocabulary panoptic segmentation performance for training on COCO and testing on ADE20K and training on ADE20K and testing on COCO. }
\label{tab:cocoade20k}
\vspace{-3 em}
\end{table*}

\vspace{-2mm}
\subsection{Results}
\label{sec:main_results}

\noindent \textbf{Open-Vocabulary Panoptic Segmentation between COCO and ADE20K:}\quad
In Table~\ref{tab:cocoade20k}, we present the zero-shot open-vocabulary panoptic segmentation results when trained on COCO and tested on ADE20K, and vice versa. Compared with existing state-of-the-art methods, our proposed PosSAM demonstrates significant improvements across all metrics. For instance, PosSAM (H) achieves a gain of +2.4 PQ over the second-best approach, FCCLIP \cite{yu2023convolutions}, under the COCO \(\rightarrow\) ADE20K setting, and an absolute gain of +4.6 PQ under the ADE20K \(\rightarrow\) COCO setting. Compared to CLIP-based methods such as MaskCLIP \cite{ding2022open}, FreeSeg \cite{qin2023freeseg}, and OPSNet \cite{chen2023open}, our scores for PQ, mAP, and mIOU show a substantial margin of improvement. This is because our method synergizes SAM for robust segmentation and CLIP for robust classification. Additionally, compared to ODISE \cite{xu2023odise}, using the Stable Diffusion UNet encoder, PosSAM outperforms this method by 20.5$\%$ and 21.8$\%$ in PQ on COCO \(\rightarrow\) ADE20K and ADE20K \(\rightarrow\) COCO respectively. Thanks to SAM's precise localization, PosSAM consistently exhibits superior segmentation quality (SQ) for Base, Large, and Huge models. Please see Appendix B for the closed set performance on both ADE20K and COCO datasets.

\noindent \textbf{Open-Vocabulary Semantic Segmentation:}\quad
In Table~\ref{tab:semantic}, we evaluate PosSAM for open-vocabulary semantic segmentation task, trained on the COCO dataset and tested across large-scale ADE20K with 150 classes (A-150) and 847 classes (A-847), as well as PASCAL-Context with 459 classes (PC-459). For all the datasets, PosSAM achieves the highest mIoU scores, indicating superior segmentation accuracy. Its mIoU scores of 14.9, 18.7, and 35.1 for A-847, PC-459, and A-150, respectively, are notably higher than those of established methods like OVSeg \cite{liang2023open}, SAN\cite{xu2023side}, ODISE \cite{xu2023open} FCCLIP\cite{yu2023convolutions} and MaskQCLIP \cite{xu2023masqclip}. This robust performance across varied class sizes and contexts highlights PosSAM's effectiveness in adapting to diverse semantic segmentation challenges and its suitability to other segmentation tasks and generalization to unseen categories, even though trained with panoptic-segmentation data only. 

\begin{table}[!ht]
\centering
\label{tab:semseg}
\small
\resizebox{0.7\linewidth}{!}{\begin{tabular}{l|ccc}
\toprule
                                    & \multicolumn{3}{c}{mIoU} \\                                                                   
Method                               & A-847         & PC-459        & A-150          \\ \hline
GroupViT~\cite{xu2022groupvit}  (CVPR 2022)     & 4.3           & 4.9           & 10.6          \\
SimBaseline~\cite{xu2021simple}  (ECCV 2022)               & -             & -             & 15.3                    \\
ZegFormer~\cite{ding2022decoupling}  (CVPR 2022)              & -             & -             & 16.4                \\
LSeg~\cite{li2022language} (ICLR 2022)                & 3.8           & 7.8           & 18.0                   \\
OVSeg~\cite{liang2023open}   (CVPR 2023)           & 9.0           & 12.4           & 29.6                      \\
SAN~\cite{xu2023side}   (CVPR 2023)            & 13.7           & 17.1           & 33.3                     \\
OpenSeg~\cite{ghiasi2022scaling} (ECCV 2022)         & 6.3           & 9.0           & 21.1             \\
MaskCLIP~\cite{ding2022open}  (ICML 2023)                & 8.2           & 10.0          & 23.7                        \\
ODISE~\cite{xu2023open} (CVPR 2023)         & 11.0 & 13.8 & 28.7   \\
MaskQCLIP~\cite{xu2023masqclip} (ICCV 2023)          & 10.7 & 18.2 & 30.4    \\
FCCLIP~\cite{yu2023convolutions}  (NeurIPS 2023)         & 14.8 & 18.2 & 34.1   \\
\midrule
\rowcolor{Gray} PosSAM         &  \textbf{14.9} &  \textbf{18.7} &  \textbf{35.1}    \\
\bottomrule
\end{tabular}}
\caption{
    \label{tab:semantic}
    {Zero-shot open-vocabulary semantic segmentation results. All methods are trained on COCO and tested on A-150 and A-847, representing ADE20K with 150 classes and 847 classes, respectively. P-459 represents PASCAL-Context with 459 classes. The best results are bolded.}
}
\vspace{-2.4 em}
\end{table}

\noindent \textbf{Effect of LDP module on PQ\textsubscript{seen} and PQ\textsubscript{unseen}:} Table~\ref{tab:novel} presents different approaches for seen and unseen panoptic segmentation in the COCO \(\rightarrow\) ADE20K setting. We observe that PosSAM outperforms the others across all metrics. Specifically, for unseen classes (PQ\(_{\text{unseen}}\)), PosSAM achieves a score of 21.3, an 18.3\% improvement over the next best method, FCCLIP~\cite{yu2023convolutions} (see Figure~\ref{fig:visual_novel}). Because FCCLIP relies solely on discriminative CLIP features, the model is biased towards the training set, i.e., seen classes. To mitigate this bias, we effectively integrate class-agnostic features from SAM leveraging the proposed LDP module. Notably, when comparing PQ\(_{\text{seen}}\) and PQ\(_{\text{unseen}}\) together, PosSAM does not overfit to training classes and generalizes well to unseen object categories. Moreover, the more specific quality metrics, PQ\(^{\text{th}}\) and PQ\(^{\text{st}}\), also achieve the highest scores with the proposed method, at 28.3 and 30.9, respectively. Overall, PQ, PQ\(^{\text{th}}\), and PQ\(^{\text{st}}\) are consistently better across all methods, demonstrating PosSAM's LDP effectiveness in not overfitting to training concepts.

\begin{table}[ht]
\vspace{- 1.5 em}
\small
\centering
\resizebox{0.8\linewidth}{!}{\begin{tabular}{lccccc}
\toprule
Method   & PQ$_{unseen}$ & PQ$_{seen}$ & PQ$^{th}$  & PQ$^{st}$ & PQ\\ \midrule
ODISE(Caption)~\cite{xu2023open}    & 15.5        & 33.2         &     21.8  &  25.9 & 23.2 \\ 
ODISE(Label)~\cite{xu2023open}    &  11.5       & 37.5         &  21.3     & 25.7  & 22.7 \\ 
FCCLIP~\cite{yu2023convolutions}   & 17.3         & 39.1         & 25.1     & 30.1  & 26.8 \\ \midrule
\rowcolor{Gray} PosSAM    &   \textbf{21.3}       &   \textbf{39.5}       &  \textbf{28.3}    &  \textbf{30.9}  &  \textbf{29.2}  \\ 
\bottomrule
\end{tabular}}
\caption{
    Comparison on Panoptic Quality (PQ) for seen and unseen classes for COCO $\rightarrow$ ADE20K. 
    Note that results of unseen classes show generalization ability of models.
}
\label{tab:novel}
\vspace{- 2.5 em}
\end{table}

\noindent \textbf{Individual Contributions of Different Components:}\quad 
In Table \ref{tab:multiple_comp}, we present study to investigate how much different components of the proposed method contribute towards its overall performance. The study begins with a baseline model, with following components progressively introduced: Feature Pyramid Network (FPN), Local Discriminative Pooling (LDP), Intersection over Union (IoU) head, and Mask-aware Selective Ensemble (MASE). Integrating FPN alone significantly boosts different performance metrics, suggesting its effectiveness in improving multi-scale feature representations. Adding LDP to FPN led to 3.4$\%$ improvements in panoptic segmentation quality due to better modeling of classification features. The integration of the IoU head, resulted in improved PQ, mAP and mIOU, since IoU helps achieve stable mask generation. Finally, all the components of our model complement each other and achieve the highest scores across all different metrics enhancing the model's ability to generate robust panoptic mask in an open-vocabulary setting. 

\begin{table}[t]
\centering
\resizebox{0.8\linewidth}{!}
{\begin{tabular}{cccc|ccccc}
\toprule
 FPN  & LDP & IoU & MASE & PQ  & $PQ^{th}$  & $PQ^{st}$  & mAP  & mIoU\\ \midrule
\rowcolor{Gray} \xmark & \xmark & \xmark & \xmark & 21.3 & 22.6 & 23.2 & 12.8  & 29.5\\
\rowcolor{Gray} \cmark & \xmark & \xmark & \xmark & 24.5 &  24.3 & 24.7  &  14.7 & 32.0\\
\rowcolor{Gray} \cmark & \cmark & \xmark & \xmark & 25.7 & 25.0  & 27.6 & 15.7  & 32.5\\
\rowcolor{Gray} \cmark & \cmark & \cmark & \xmark & 25.9 & 25.2 & 27.9 & 16.0 & \textbf{33.0} \\
\rowcolor{Gray} \cmark & \cmark & \cmark & \cmark & \textbf{26.5} & \textbf{25.5} &  \textbf{28.6} & \textbf{16.8}  & 32.9\\
\bottomrule
\end{tabular}}
\caption{Ablation study on the proposed components for PosSAM (L), performed by training on COCO and validating on ADE20K. FPN: Feature Pyramid Network, LDP: Local Discriminative Pooling, IoU: IoU Loss, MASE - Mask-aware Selective Ensemble}
\label{tab:multiple_comp}
\vspace{- 1 em}
\end{table}

\noindent \textbf{Effect of Varying the Pooling Strategy:}\quad 
Table~\ref{tab:pool} compares different pooling mechanisms and
impact on 
 PQ, mIoU, and mAP when employing SAM \cite{kirillov2023segment} and CLIP \cite{radford2021learning} features in the COCO \(\rightarrow\) ADE20K setting. We observe that Mask Pooling with SAM features establishes a baseline PQ of 28.0, a marginal improvement over CLIP features, indicating the discriminative advantage of CLIP. In the Cross-Attention (Cross-Attn) approach, we treat SAM features as the query and CLIP features as the key and value. For Cross-Attn, the incremental improvements in PQ  and mIoU suggest that using both features results in enhanced classification. Finally, Local Discriminative Pooling (LDP) better combines SAM and CLIP features, leading to more robust classification and achieving the highest scores across all metrics. Notably, the increase in PQ to 29.2, demonstrating its effectiveness in the aggregation of SAM and CLIP embeddings. This suggests that our proposed LDP module leverages complementary strengths of SAM and CLIP for improved open-vocabulary segmentation.

\begin{table}[t]
\centering
\resizebox{0.8\linewidth}{!}{\begin{tabular}{l|c|ccc}
\toprule
Pooling  & Features & PQ   & mIoU  & mAP\\
\midrule 
\rowcolor{Gray} Mask Pooling & SAM Feat & 28.0 &  33.7 & 17.7 \\
\rowcolor{Gray} Mask Pooling & CLIP Feat & 26.4 & 33.3  & 16.8 \\
\rowcolor{Gray} Cross-Attn & SAM Feat, CLIP Feat & 28.3 &  34.0 & 17.5 \\
\rowcolor{Gray} Local Disc Pooling & SAM Feat, CLIP Feat & \textbf{29.2} & \textbf{35.1}  & \textbf{18.3} \\
\bottomrule
\end{tabular}}
\caption{Studying the performance obtained by varying our pooling approach for PosSAM (H) variants trained on COCO and inference on ADE20K.}
\label{tab:pool}
\vspace{- 2.5 em}
\end{table}

\begin{table}[ht]
\centering
\label{tab:component}
\resizebox{0.7\linewidth}{!}{
\begin{tabular}{lcccc}
\toprule
Pooling  & No. of. Heads & PQ   & $PQ_{th}$  & $PQ_{st}$\\
\hline
\rowcolor{Gray} Two-way Decoder & 1 & 12.0 &  5.0 & 28.6 \\
\rowcolor{Gray} Two-way Decoder & 9 & 14.2 & 5.8  & 28.7\\
\rowcolor{Gray} Mask2Former & 9 &  \textbf{26.5} & \textbf{16.8}  & \textbf{32.9}\\
\bottomrule
\end{tabular}}
\caption{Ablation analysis on the SAM (L) Two-way decoder and Mask2former decoder for panoptic mask generation, trained on COCO and inference on ADE20K.}
\label{tab:decoder}
\end{table}

\noindent \textbf{Comparison of SAM and Mask2former Decoders:}\quad 
The ablation study presented in the Table \ref{tab:decoder} focuses on evaluating the performance of different decoder variations on panoptic mask generation for COCO $\rightarrow$ ADE20K setting. The study compares a SAM's two-way decoder with a single head against the same decoder with nine heads, and also contrasts these with the Mask2Former decoder \cite{cheng2021mask2former}. The single head two-way decoder demonstrates the lowest performance across all metrics and increasing the number of heads to nine shows a reasonable improvement. However, the Mask2Former decoder exhibits a significant performance boost, achieving the highest scores (\(PQ: 26.5\), \(PQ_{\text{th}}: 16.8\), \(PQ_{\text{st}}: 32.9\)), indicating that the Mask2Former architecture with multi-resolution setup and deformable attention is more effective for panoptic mask generation. Furthermore, the SAM decoder is designed to be lightweight with weaker attention, so that the encoder is forced to do most of the segmentation in its feature space itself (see Appendix B for more details).

\noindent \textbf{Effect of Varying the Ensembling Strategy during inference:}\quad 
Table~\ref{tab:ensemble} compares different ensembling methods during inference. By varying $\alpha$ and $\beta$ to vary reliance on the individual encodings, we demonstrate the effect of each component. Geometric ensembling, as proposed by~\cite{xu2023odise}, is an effective way to combine the two embeddings. However, we observe that using our MASE strategy boosts the PQ by 2.8\%.

\begin{table}[ht]
\vspace{-2mm}
\centering
\resizebox{0.8\linewidth}{!}{\begin{tabular}{cc|ccc|c}
\toprule
$\alpha$ & $\beta$ & Seen Class & Unseen Class & Ensembling & PQ \\
\midrule
 \rowcolor{Gray} 0.0 &  0.0 & $\mathbf{G_{ldp}}$ & $\mathbf{G_{ldp}}$ & None & 23.4 \\
 \rowcolor{Gray} 1.0 & 1.0 & $\mathbf{G_{clip}}$ & $\mathbf{G_{clip}}$ & None & 25.6 \\
 \rowcolor{Gray} 0.4 &  0.8 & Ensemble & Ensemble & Geometric & 28.4 \\
\rowcolor{Gray} 0.4 &  0.8 & Ensemble & Ensemble & MASE & \textbf{29.2} \\
\bottomrule
\end{tabular}}
\caption{Effect of varying the ensembling strategy for seen and unseen classes during inference, for a PosSAM (H) variant trained on COCO and inference on ADE20K.}
\label{tab:ensemble}
\end{table}

\section{Conclusion}
In conclusion, this work introduces PosSAM, an open-vocabulary panoptic segmentation framework that leverages SAM by addressing its limitations to generate class and instance-aware masks. Through comprehensive experiments, we demonstrate how the novel Local Discriminative Pooling (LDP) module significantly reduces bias towards seen classes, thereby significantly improving unseen open-vocabulary classification. Moreover, our Mask-Aware Selective Ensembling (MASE) algorithm employs an adaptive strategy to accurately differentiate between seen and unseen classes by leveraging both IoU and LDP confidence scores during inference. Our model achieves state-of-the-art performance across various OV benchmark datasets, highlighting PosSAM's effectiveness and efficiency.


%
%
\bibliographystyle{splncs04}
\bibliography{main}

\begin{thebibliography}{10}
\providecommand{\url}[1]{\texttt{#1}}
\providecommand{\urlprefix}{URL }
\providecommand{\doi}[1]{https://doi.org/#1}

\bibitem{alayrac2022flamingo}
Alayrac, J.B., Donahue, J., Luc, P., Miech, A., Barr, I., Hasson, Y., Lenc, K., Mensch, A., Millican, K., Reynolds, M., et~al.: Flamingo: a visual language model for few-shot learning. In: NeurIPS (2022)

\bibitem{borse2021hs3}
Borse, S., Cai, H., Zhang, Y., Porikli, F.: Hs3: Learning with proper task complexity in hierarchically supervised semantic segmentation. arXiv preprint arXiv:2111.02333  (2021)

\bibitem{borse2023dejavu}
Borse, S., Das, D., Park, H., Cai, H., Garrepalli, R., Porikli, F.: Dejavu: Conditional regenerative learning to enhance dense prediction. In: Proceedings of the IEEE/CVF Conference on Computer Vision and Pattern Recognition. pp. 19466--19477 (2023)

\bibitem{borse2023x}
Borse, S., Klingner, M., Kumar, V.R., Cai, H., Almuzairee, A., Yogamani, S., Porikli, F.: X-align: Cross-modal cross-view alignment for bird's-eye-view segmentation. In: Proceedings of the IEEE/CVF Winter Conference on Applications of Computer Vision. pp. 3287--3297 (2023)

\bibitem{borse2022panoptic}
Borse, S., Park, H., Cai, H., Das, D., Garrepalli, R., Porikli, F.: Panoptic, instance and semantic relations: A relational context encoder to enhance panoptic segmentation. In: Proceedings of the IEEE/CVF Conference on Computer Vision and Pattern Recognition. pp. 1269--1279 (2022)

\bibitem{borse2021inverseform}
Borse, S., Wang, Y., Zhang, Y., Porikli, F.: Inverseform: A loss function for structured boundary-aware segmentation. In: Proceedings of the IEEE/CVF Conference on Computer Vision and Pattern Recognition. pp. 5901--5911 (2021)

\bibitem{carion2020end}
Carion, N., Massa, F., Synnaeve, G., Usunier, N., Kirillov, A., Zagoruyko, S.: End-to-end object detection with transformers. In: European conference on computer vision. pp. 213--229. Springer (2020)

\bibitem{chen2023semantic}
Chen, J., Yang, Z., Zhang, L.: Semantic segment anything. \url{https://github.com/fudan-zvg/Semantic-Segment-Anything} (2023)

\bibitem{deeplabV2}
Chen, L.C., Papandreou, G., Kokkinos, I., Murphy, K., Yuille, A.L.: {DeepLab}: Semantic image segmentation with deep convolutional nets, atrous convolution, and fully connected {CRF}s. IEEE TPAMI  (2018)

\bibitem{deeplabV3}
Chen, L.C., Papandreou, G., Schroff, F., Adam, H.: Rethinking atrous convolution for semantic image segmentation. arXiv:1706.05587  (2017)

\bibitem{chen2023open}
Chen, X., Li, S., Lim, S.N., Torralba, A., Zhao, H.: Open-vocabulary panoptic segmentation with embedding modulation. arXiv preprint arXiv:2303.11324  (2023)

\bibitem{cheng2022masked}
Cheng, B., Misra, I., Schwing, A.G., Kirillov, A., Girdhar, R.: Masked-attention mask transformer for universal image segmentation. In: Proceedings of the IEEE/CVF conference on computer vision and pattern recognition. pp. 1290--1299 (2022)

\bibitem{cheng2021masked}
Cheng, B., Misra, I., Schwing, A.G., Kirillov, A., Girdhar, R.: Masked-attention mask transformer for universal image segmentation. In: CVPR (2022)

\bibitem{cheng2021mask2former}
Cheng, B., Misra, I., Schwing, A.G., Kirillov, A., Girdhar, R.: Masked-attention mask transformer for universal image segmentation (2022)

\bibitem{cheng2023segment}
Cheng, Y., Li, L., Xu, Y., Li, X., Yang, Z., Wang, W., Yang, Y.: Segment and track anything. arXiv preprint arXiv:2305.06558  (2023)

\bibitem{cherti2023reproducible}
Cherti, M., Beaumont, R., Wightman, R., Wortsman, M., Ilharco, G., Gordon, C., Schuhmann, C., Schmidt, L., Jitsev, J.: Reproducible scaling laws for contrastive language-image learning. In: Proceedings of the IEEE/CVF Conference on Computer Vision and Pattern Recognition. pp. 2818--2829 (2023)

\bibitem{ding2022decoupling}
Ding, J., Xue, N., Xia, G.S., Dai, D.: Decoupling zero-shot semantic segmentation. In: Proceedings of the IEEE/CVF Conference on Computer Vision and Pattern Recognition. pp. 11583--11592 (2022)

\bibitem{ding2022open}
Ding, Z., Wang, J., Tu, Z.: Open-vocabulary universal image segmentation with maskclip. In: {ICML} (2023)

\bibitem{dosovitskiy2020image}
Dosovitskiy, A., Beyer, L., Kolesnikov, A., Weissenborn, D., Zhai, X., Unterthiner, T., Dehghani, M., Minderer, M., Heigold, G., Gelly, S., et~al.: An image is worth 16x16 words: Transformers for image recognition at scale. In: ICLR (2021)

\bibitem{everingham2010pascal}
Everingham, M., Van~Gool, L., Williams, C.K., Winn, J., Zisserman, A.: The pascal visual object classes (voc) challenge. IJCV  \textbf{88},  303--338 (2010)

\bibitem{furst2022cloob}
F{\"u}rst, A., Rumetshofer, E., Lehner, J., Tran, V.T., Tang, F., Ramsauer, H., Kreil, D., Kopp, M., Klambauer, G., Bitto, A., et~al.: Cloob: Modern hopfield networks with infoloob outperform clip. Advances in neural information processing systems  \textbf{35},  20450--20468 (2022)

\bibitem{ghiasi2021simple}
Ghiasi, G., Cui, Y., Srinivas, A., Qian, R., Lin, T.Y., Cubuk, E.D., Le, Q.V., Zoph, B.: Simple copy-paste is a strong data augmentation method for instance segmentation. In: CVPR (2021)

\bibitem{ghiasi2022scaling}
Ghiasi, G., Gu, X., Cui, Y., Lin, T.Y.: Scaling open-vocabulary image segmentation with image-level labels. In: European Conference on Computer Vision. pp. 540--557. Springer (2022)

\bibitem{gu2021open}
Gu, X., Lin, T.Y., Kuo, W., Cui, Y.: Open-vocabulary object detection via vision and language knowledge distillation. In: ICLR (2022)

\bibitem{gupta2020contrastive}
Gupta, T., Vahdat, A., Chechik, G., Yang, X., Kautz, J., Hoiem, D.: Contrastive learning for weakly supervised phrase grounding. In: ECCV (2020)

\bibitem{han2023segment}
Han, D., Zhang, C., Qiao, Y., Qamar, M., Jung, Y., Lee, S., Bae, S.H., Hong, C.S.: Segment anything model (sam) meets glass: Mirror and transparent objects cannot be easily detected. arXiv preprint arXiv:2305.00278  (2023)

\bibitem{huang2019ccnet}
Huang, Z., Wang, X., Huang, L., Huang, C., Wei, Y., Liu, W.: Ccnet: Criss-cross attention for semantic segmentation. In: ICCV (2019)

\bibitem{groundedsam}
IDEA-Research: {Grounded-Segment-Anything}. \url{https://github.com/IDEA-Research/Grounded-Segment-Anything} (2023)

\bibitem{ilharco_gabriel_2021_5143773}
Ilharco, G., Wortsman, M., Wightman, R., Gordon, C., Carlini, N., Taori, R., Dave, A., Shankar, V., Namkoong, H., Miller, J., Hajishirzi, H., Farhadi, A., Schmidt, L.: Openclip (Jul 2021). \doi{10.5281/zenodo.5143773}, \url{https://doi.org/10.5281/zenodo.5143773}, if you use this software, please cite it as below.

\bibitem{jia2021scaling}
Jia, C., Yang, Y., Xia, Y., Chen, Y.T., Parekh, Z., Pham, H., Le, Q., Sung, Y.H., Li, Z., Duerig, T.: Scaling up visual and vision-language representation learning with noisy text supervision. In: International conference on machine learning. pp. 4904--4916. PMLR (2021)

\bibitem{kang2022any}
Kang, M., Min, D., Hwang, S.J.: Any-speaker adaptive text-to-speech synthesis with diffusion models. arXiv preprint arXiv:2211.09383  (2022)

\bibitem{kingma2014adam}
Kingma, D.P., Ba, J.: Adam: A method for stochastic optimization. In: ICLR (2015)

\bibitem{kirillov2018panoptic}
Kirillov, A., He, K., Girshick, R., Rother, C., Doll{\'a}r, P.: Panoptic segmentation. In: CVPR (2019)

\bibitem{kirillov2023segment}
Kirillov, A., Mintun, E., Ravi, N., Mao, H., Rolland, C., Gustafson, L., Xiao, T., Whitehead, S., Berg, A.C., Lo, W.Y., et~al.: Segment anything. arXiv preprint arXiv:2304.02643  (2023)

\bibitem{li2022languagedriven}
Li, B., Weinberger, K.Q., Belongie, S., Koltun, V., Ranftl, R.: Language-driven semantic segmentation. In: International Conference on Learning Representations (2022), \url{https://openreview.net/forum?id=RriDjddCLN}

\bibitem{li2022language}
Li, B., Weinberger, K.Q., Belongie, S., Koltun, V., Ranftl, R.: Language-driven semantic segmentation. In: ICLR (2022)

\bibitem{li2023mask}
Li, F., Zhang, H., Liu, S., Zhang, L., Ni, L.M., Shum, H.Y., et~al.: Mask dino: Towards a unified transformer-based framework for object detection and segmentation. In: CVPR (2023)

\bibitem{li2023semantic}
Li, F., Zhang, H., Sun, P., Zou, X., Liu, S., Yang, J., Li, C., Zhang, L., Gao, J.: Semantic-sam: Segment and recognize anything at any granularity. arXiv preprint arXiv:2307.04767  (2023)

\bibitem{li2022blip}
Li, J., Li, D., Xiong, C., Hoi, S.: Blip: Bootstrapping language-image pre-training for unified vision-language understanding and generation. In: International Conference on Machine Learning. pp. 12888--12900. PMLR (2022)

\bibitem{li2019visualbert}
Li, L.H., Yatskar, M., Yin, D., Hsieh, C.J., Chang, K.W.: Visualbert: A simple and performant baseline for vision and language. arXiv preprint arXiv:1908.03557  (2019)

\bibitem{li2024omg}
Li, X., Yuan, H., Li, W., Ding, H., Wu, S., Zhang, W., Li, Y., Chen, K., Loy, C.C.: Omg-seg: Is one model good enough for all segmentation? arXiv preprint arXiv:2401.10229  (2024)

\bibitem{li2021supervision}
Li, Y., Liang, F., Zhao, L., Cui, Y., Ouyang, W., Shao, J., Yu, F., Yan, J.: Supervision exists everywhere: A data efficient contrastive language-image pre-training paradigm. arXiv preprint arXiv:2110.05208  (2021)

\bibitem{li2022exploring}
Li, Y., Mao, H., Girshick, R., He, K.: Exploring plain vision transformer backbones for object detection. In: ECCV (2022)

\bibitem{liang2023open}
Liang, F., Wu, B., Dai, X., Li, K., Zhao, Y., Zhang, H., Zhang, P., Vajda, P., Marculescu, D.: Open-vocabulary semantic segmentation with mask-adapted clip. In: Proceedings of the IEEE/CVF Conference on Computer Vision and Pattern Recognition. pp. 7061--7070 (2023)

\bibitem{lin2014microsoft}
Lin, T.Y., Maire, M., Belongie, S., Hays, J., Perona, P., Ramanan, D., Doll{\'a}r, P., Zitnick, C.L.: Microsoft coco: Common objects in context. In: Computer Vision--ECCV 2014: 13th European Conference, Zurich, Switzerland, September 6-12, 2014, Proceedings, Part V 13. pp. 740--755. Springer (2014)

\bibitem{liu2023grounding}
Liu, S., Zeng, Z., Ren, T., Li, F., Zhang, H., Yang, J., Li, C., Yang, J., Su, H., Zhu, J., et~al.: Grounding dino: Marrying dino with grounded pre-training for open-set object detection. arXiv preprint arXiv:2303.05499  (2023)

\bibitem{long2015fully}
Long, J., Shelhamer, E., Darrell, T.: Fully convolutional networks for semantic segmentation. In: CVPR (2015)

\bibitem{loshchilov2017decoupled}
Loshchilov, I., Hutter, F.: Decoupled weight decay regularization. In: ICLR (2019)

\bibitem{lu2019vilbert}
Lu, J., Batra, D., Parikh, D., Lee, S.: Vilbert: Pretraining task-agnostic visiolinguistic representations for vision-and-language tasks. Advances in neural information processing systems  \textbf{32} (2019)

\bibitem{ma2022open}
Ma, C., Yang, Y., Wang, Y., Zhang, Y., Xie, W.: Open-vocabulary semantic segmentation with frozen vision-language models. In: BMVC (2022)

\bibitem{ma2023segment}
Ma, J., Wang, B.: Segment anything in medical images. arXiv preprint arXiv:2304.12306  (2023)

\bibitem{manas2022mapl}
Ma{\~n}as, O., Rodriguez, P., Ahmadi, S., Nematzadeh, A., Goyal, Y., Agrawal, A.: Mapl: Parameter-efficient adaptation of unimodal pre-trained models for vision-language few-shot prompting. arXiv preprint arXiv:2210.07179  (2022)

\bibitem{mottaghi2014role}
Mottaghi, R., Chen, X., Liu, X., Cho, N.G., Lee, S.W., Fidler, S., Urtasun, R., Yuille, A.: The role of context for object detection and semantic segmentation in the wild. In: CVPR (2014)

\bibitem{park2023segclip}
Park, C.: Segment anything with clip. \url{https://github.com/Curt-Park/segment-anything-with-clip} (2023)

\bibitem{qin2023freeseg}
Qin, J., Wu, J., Yan, P., Li, M., Yuxi, R., Xiao, X., Wang, Y., Wang, R., Wen, S., Pan, X., et~al.: Freeseg: Unified, universal and open-vocabulary image segmentation. In: Proceedings of the IEEE/CVF Conference on Computer Vision and Pattern Recognition. pp. 19446--19455 (2023)

\bibitem{radford2021learning}
Radford, A., Kim, J.W., Hallacy, C., Ramesh, A., Goh, G., Agarwal, S., Sastry, G., Askell, A., Mishkin, P., Clark, J., et~al.: Learning transferable visual models from natural language supervision. In: International conference on machine learning. pp. 8748--8763. PMLR (2021)

\bibitem{rombach2022high}
Rombach, R., Blattmann, A., Lorenz, D., Esser, P., Ommer, B.: High-resolution image synthesis with latent diffusion models. In: Proceedings of the IEEE/CVF conference on computer vision and pattern recognition. pp. 10684--10695 (2022)

\bibitem{shen2023anything}
Shen, Q., Yang, X., Wang, X.: Anything-3d: Towards single-view anything reconstruction in the wild. arXiv preprint arXiv:2304.10261  (2023)

\bibitem{singh2022flava}
Singh, A., Hu, R., Goswami, V., Couairon, G., Galuba, W., Rohrbach, M., Kiela, D.: Flava: A foundational language and vision alignment model. In: Proceedings of the IEEE/CVF Conference on Computer Vision and Pattern Recognition. pp. 15638--15650 (2022)

\bibitem{strudel2021segmenter}
Strudel, R., Garcia, R., Laptev, I., Schmid, C.: Segmenter: Transformer for semantic segmentation. In: ICCV (2021)

\bibitem{tan2019lxmert}
Tan, H., Bansal, M.: Lxmert: Learning cross-modality encoder representations from transformers. In: EMNLP (2019)

\bibitem{tang2023can}
Tang, L., Xiao, H., Li, B.: Can sam segment anything? when sam meets camouflaged object detection. arXiv preprint arXiv:2304.04709  (2023)

\bibitem{tariq2023segment}
Tariq, S., Arfeto, B.E., Zhang, C., Shin, H.: Segment anything meets semantic communication. arXiv preprint arXiv:2306.02094  (2023)

\bibitem{tsimpoukelli2021multimodal}
Tsimpoukelli, M., Menick, J.L., Cabi, S., Eslami, S., Vinyals, O., Hill, F.: Multimodal few-shot learning with frozen language models. Advances in Neural Information Processing Systems  \textbf{34},  200--212 (2021)

\bibitem{vs2023instance}
VS, V., Oza, P., Patel, V.M.: Instance relation graph guided source-free domain adaptive object detection. In: Proceedings of the IEEE/CVF Conference on Computer Vision and Pattern Recognition. pp. 3520--3530 (2023)

\bibitem{vs2023towards}
VS, V., Oza, P., Patel, V.M.: Towards online domain adaptive object detection. In: Proceedings of the IEEE/CVF Winter Conference on Applications of Computer Vision. pp. 478--488 (2023)

\bibitem{vs2023mask}
VS, V., Yu, N., Xing, C., Qin, C., Gao, M., Niebles, J.C., Patel, V.M., Xu, R.: Mask-free ovis: Open-vocabulary instance segmentation without manual mask annotations. In: Proceedings of the IEEE/CVF Conference on Computer Vision and Pattern Recognition. pp. 23539--23549 (2023)

\bibitem{wang2021simvlm}
Wang, Z., Yu, J., Yu, A.W., Dai, Z., Tsvetkov, Y., Cao, Y.: Simvlm: Simple visual language model pretraining with weak supervision. arXiv preprint arXiv:2108.10904  (2021)

\bibitem{wu2022tinyvit}
Wu, K., Zhang, J., Peng, H., Liu, M., Xiao, B., Fu, J., Yuan, L.: Tinyvit: Fast pretraining distillation for small vision transformers. In: European Conference on Computer Vision. pp. 68--85. Springer (2022)

\bibitem{xu2022groupvit}
Xu, J., De~Mello, S., Liu, S., Byeon, W., Breuel, T., Kautz, J., Wang, X.: Groupvit: Semantic segmentation emerges from text supervision. In: Proceedings of the IEEE/CVF Conference on Computer Vision and Pattern Recognition. pp. 18134--18144 (2022)

\bibitem{xu2023odise}
Xu, J., Liu, S., Vahdat, A., Byeon, W., Wang, X., De~Mello, S.: {Open-Vocabulary Panoptic Segmentation with Text-to-Image Diffusion Models}. arXiv preprint arXiv:2303.04803  (2023)

\bibitem{xu2023open}
Xu, J., Liu, S., Vahdat, A., Byeon, W., Wang, X., De~Mello, S.: Open-vocabulary panoptic segmentation with text-to-image diffusion models. In: Proceedings of the IEEE/CVF Conference on Computer Vision and Pattern Recognition. pp. 2955--2966 (2023)

\bibitem{xu2023side}
Xu, M., Zhang, Z., Wei, F., Hu, H., Bai, X.: Side adapter network for open-vocabulary semantic segmentation. In: Proceedings of the IEEE/CVF Conference on Computer Vision and Pattern Recognition. pp. 2945--2954 (2023)

\bibitem{xu2021simple}
Xu, M., Zhang, Z., Wei, F., Lin, Y., Cao, Y., Hu, H., Bai, X.: A simple baseline for zero-shot semantic segmentation with pre-trained vision-language model. In: ECCV (2022)

\bibitem{xu2023bridgetower}
Xu, X., Wu, C., Rosenman, S., Lal, V., Che, W., Duan, N.: Bridgetower: Building bridges between encoders in vision-language representation learning. In: Proceedings of the AAAI Conference on Artificial Intelligence. vol.~37, pp. 10637--10647 (2023)

\bibitem{xu2023masqclip}
Xu, X., Xiong, T., Ding, Z., Tu, Z.: Masqclip for open-vocabulary universal image segmentation. In: Proceedings of the IEEE/CVF International Conference on Computer Vision. pp. 887--898 (2023)

\bibitem{yang2023track}
Yang, J., Gao, M., Li, Z., Gao, S., Wang, F., Zheng, F.: Track anything: Segment anything meets videos. arXiv preprint arXiv:2304.11968  (2023)

\bibitem{yu2023convolutions}
Yu, Q., He, J., Deng, X., Shen, X., Chen, L.C.: Convolutions die hard: Open-vocabulary segmentation with single frozen convolutional clip. arXiv preprint arXiv:2308.02487  (2023)

\bibitem{yu2023inpaint}
Yu, T., Feng, R., Feng, R., Liu, J., Jin, X., Zeng, W., Chen, Z.: Inpaint anything: Segment anything meets image inpainting. arXiv preprint arXiv:2304.06790  (2023)

\bibitem{yuan2018ocnet}
Yuan, Y., Huang, L., Guo, J., Zhang, C., Chen, X., Wang, J.: {OCNet}: Object context for semantic segmentation. IJCV  (2021)

\bibitem{zhai2022lit}
Zhai, X., Wang, X., Mustafa, B., Steiner, A., Keysers, D., Kolesnikov, A., Beyer, L.: Lit: Zero-shot transfer with locked-image text tuning. In: Proceedings of the IEEE/CVF Conference on Computer Vision and Pattern Recognition. pp. 18123--18133 (2022)

\bibitem{zhang2023faster}
Zhang, C., Han, D., Qiao, Y., Kim, J.U., Bae, S.H., Lee, S., Hong, C.S.: Faster segment anything: Towards lightweight sam for mobile applications. arXiv preprint arXiv:2306.14289  (2023)

\bibitem{zhang2023input}
Zhang, Y., Zhou, T., Liang, P., Chen, D.Z.: Input augmentation with sam: Boosting medical image segmentation with segmentation foundation model. arXiv preprint arXiv:2304.11332  (2023)

\bibitem{zhao2017pspnet}
Zhao, H., Shi, J., Qi, X., Wang, X., Jia, J.: Pyramid scene parsing network. In: CVPR (2017)

\bibitem{zhou2017scene}
Zhou, B., Zhao, H., Puig, X., Fidler, S., Barriuso, A., Torralba, A.: Scene parsing through ade20k dataset. In: Proceedings of the IEEE conference on computer vision and pattern recognition. pp. 633--641 (2017)

\bibitem{zhou2022extract}
Zhou, C., Loy, C.C., Dai, B.: Extract free dense labels from clip. In: European Conference on Computer Vision. pp. 696--712. Springer (2022)

\bibitem{zhou2023zegclip}
Zhou, Z., Lei, Y., Zhang, B., Liu, L., Liu, Y.: Zegclip: Towards adapting clip for zero-shot semantic segmentation. In: Proceedings of the IEEE/CVF Conference on Computer Vision and Pattern Recognition. pp. 11175--11185 (2023)

\bibitem{zou2023generalized}
Zou, X., Dou, Z.Y., Yang, J., Gan, Z., Li, L., Li, C., Dai, X., Behl, H., Wang, J., Yuan, L., et~al.: Generalized decoding for pixel, image, and language. In: Proceedings of the IEEE/CVF Conference on Computer Vision and Pattern Recognition. pp. 15116--15127 (2023)

\end{thebibliography}

\title{Appendices}
\author{}
\institute{}
\maketitle
\appendix
\counterwithin{figure}{section}
\counterwithin{table}{section}


\noindent \large{\textbf{Overview}}
\label{sec:SuppleIntro}

\noindent  As part of the supplementary materials, we present our additional Implementation Details (Section. \ref{sec:suppli_impl}), Experiments and Analysis (Section. \ref{sec:suppli_exp}) and Open-Vocabulary Visualization (Section. \ref{sec:suppli_vis}), and Limitations (Section. \ref{sec:suppli_limit}) as an extension to the ones shown in the paper.

\section{Implementation Details}
\label{sec:suppli_impl}
For a fair-comparison with recent works, we follow ODISE [1] and FCCLIP [2] and trained PosSAM for 250K iterations with image crops of size \( 1024 \times 1024 \), resizing the shorter side to \( 1024 \). We applied data augmentation techniques as proposed by \cite{ghiasi2021simple}, which included random scaling between 0.1 and 2.0. We trained our using 8 NVIDIA A100 GPUs, processing 2 images per GPU, for an effective batch size of 16. We employed the AdamW optimizer \cite{loshchilov2017decoupled} with a learning rate of \( 1 \times 10^{-4} \) and a weight decay of \( 0.05 \). The learning rate was scheduled to reduce by a factor of 10 at the 150,000th and 200,000th iterations. We utilized a ConvNext-L backbone for the CLIP architecture \cite{ilharco_gabriel_2021_5143773}. Following \cite{cheng2021masked} and \cite{carion2020end}, Hungarian matching was used to match the predicted masks with ground truth masks, and training losses were computed based on these pairs. Following \cite{cheng2021masked}, classification loss weight (\( \gamma_{a} \)) and mask loss weight (\( \gamma_{b} \)) were set to 2.0 and 5.0, respectively, with the IoU loss weight (\( \gamma_{c} \)) at 1.0. We set $\alpha=0.8$ and $\beta=0.4$ across all datasets.

\section{Additional Experiments}
\label{sec:suppli_exp}

\begin{figure}
    \centering
    \includegraphics[width=1.0\linewidth]{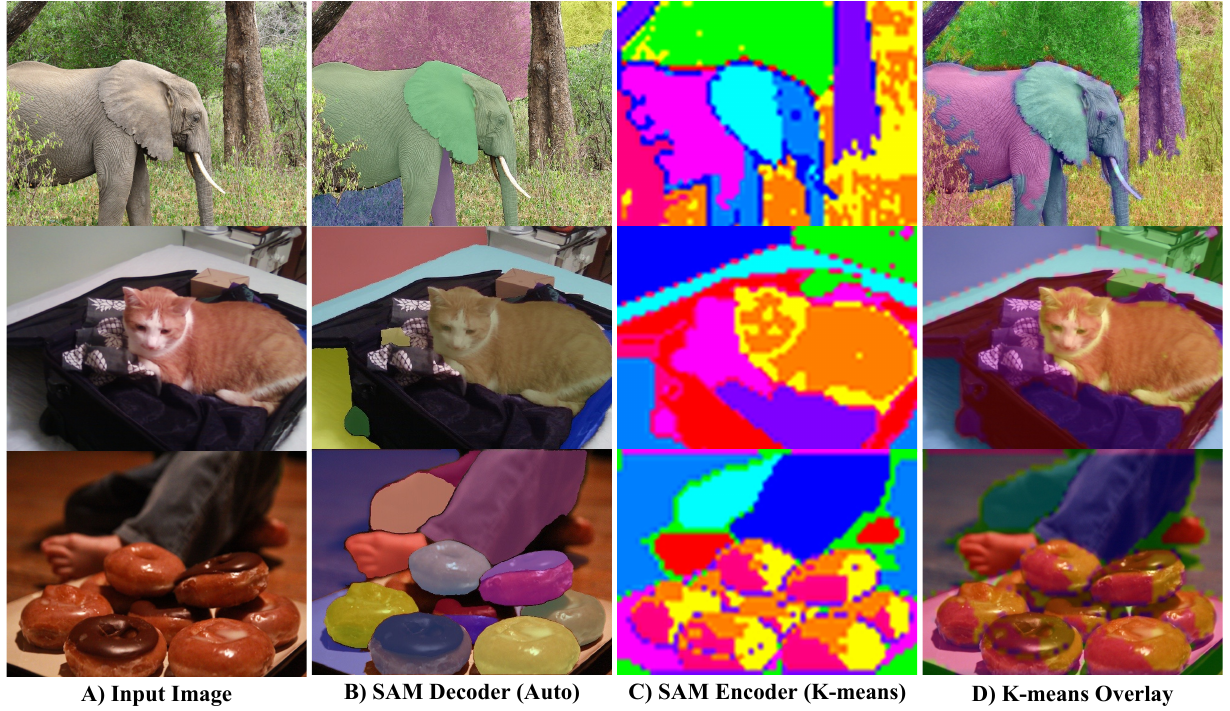}
    \vspace{- 1.0 em}
    \caption{SAM decoder vs SAM encoder K-means Clustering}
    \label{fig:enc_dec}
\end{figure}

\noindent \textbf{SAM Encoder vs Decoder:}\quad 
As shown in Fig.~\ref{fig:enc_dec}, we present a comparison between the SAM encoder features clustering mask and SAM decoder mask to verify the effectiveness of both components. Specifically, we argue that the SAM decoder is designed to be lightweight and incorporates less complex attention mechanisms, enabling the SAM encoder to undertake the majority of the segmentation task. To verify this, we feed an image and perform k-means clustering on the features from the frozen SAM encoder. For the decoder, we employ an automatic point generator that samples points uniformly across the image, with each point acting as a prompt to SAM decoder for generating masks. The clustering maps obtained from the SAM encoder features demonstrate that the majority of object segmentation has already occurred at the encoder stage, and the SAM decoder subsequently refines these features to produce a finer segmentation mask. Thus, our experiments have effectively verified that the encoder primarily learns to perform segmentation, while the decoder refines these features to achieve more detailed masks.

\begin{table}[ht]
\centering
\label{tab:component}
\resizebox{1.00\linewidth}{!}{\begin{tabular}{lccccccc}
\toprule
Model  & FPS & Train Param  & Frozen Param & Total Param  & PQ & mAP & mIOU\\
\hline
ODISE~\cite{xu2023open} & 1.26 & 28.1M &  1493.8M & 1521.9M & 23.2 & 14.0 & 28.4 \\
PosSAM (B) & 1.72 & 51.5M & 441.4M  & 493.0M & 26.5 & 16.8 & 32.9 \\
PosSAM (L) & 1.48 & 51.5M & 660.0M  & 711.6M & 28.0 & 18.3 & 34.2 \\
PosSAM (H) & 1.30 & 51.5M & 988.7M  & 1040.3M & 29.2 & 18.3 & 35.1 \\
\bottomrule
\end{tabular}}
\caption{{Parameter and Speed Analysis} for the proposed method includes metrics such as \textbf{FPS} (Frames Per Second), \textbf{Train Params} (Total parameters trained during training), \textbf{Frozen Params} (Total parameters which are frozen during training), and \textbf{Total Params} (Total parameters of the model).}
\label{tab:param_speed}
\end{table}

\noindent \textbf{Comparison of parameters:}\quad 
The Table~\ref{tab:param_speed} compares FPS, trainable and total parameters for various models, including ODISE~\cite{xu2023open}, PosSAM (B), PosSAM (L), and PosSAM (H). As shown in Table~\ref{tab:param_speed}, the trainable learnable parameters for PosSAM Base, Large, and Huge models are fixed, as all models produce 256 output channels, and the learnable parameters are FPN, mask decoder, and LDP module, which together comprise just 51.5 million parameters. As a result, the total parameters increase as the SAM backbone size increases, suggesting that less than 10\% of the model weights are being trained. Compared to ODISE, which has 1521.9 million parameters, all our model sizes are smaller but offer higher FPS and a significant boost in performance across all metrics. In summary, while ODISE is slower with many frozen parameters, PosSAM models balance FPS and parameter counts, with PosSAM (B) being the fastest and PosSAM (H) having the most parameters with overall better performance for all PosSAM variations.

\begin{figure}
    \centering
    \includegraphics[width=0.9\linewidth]{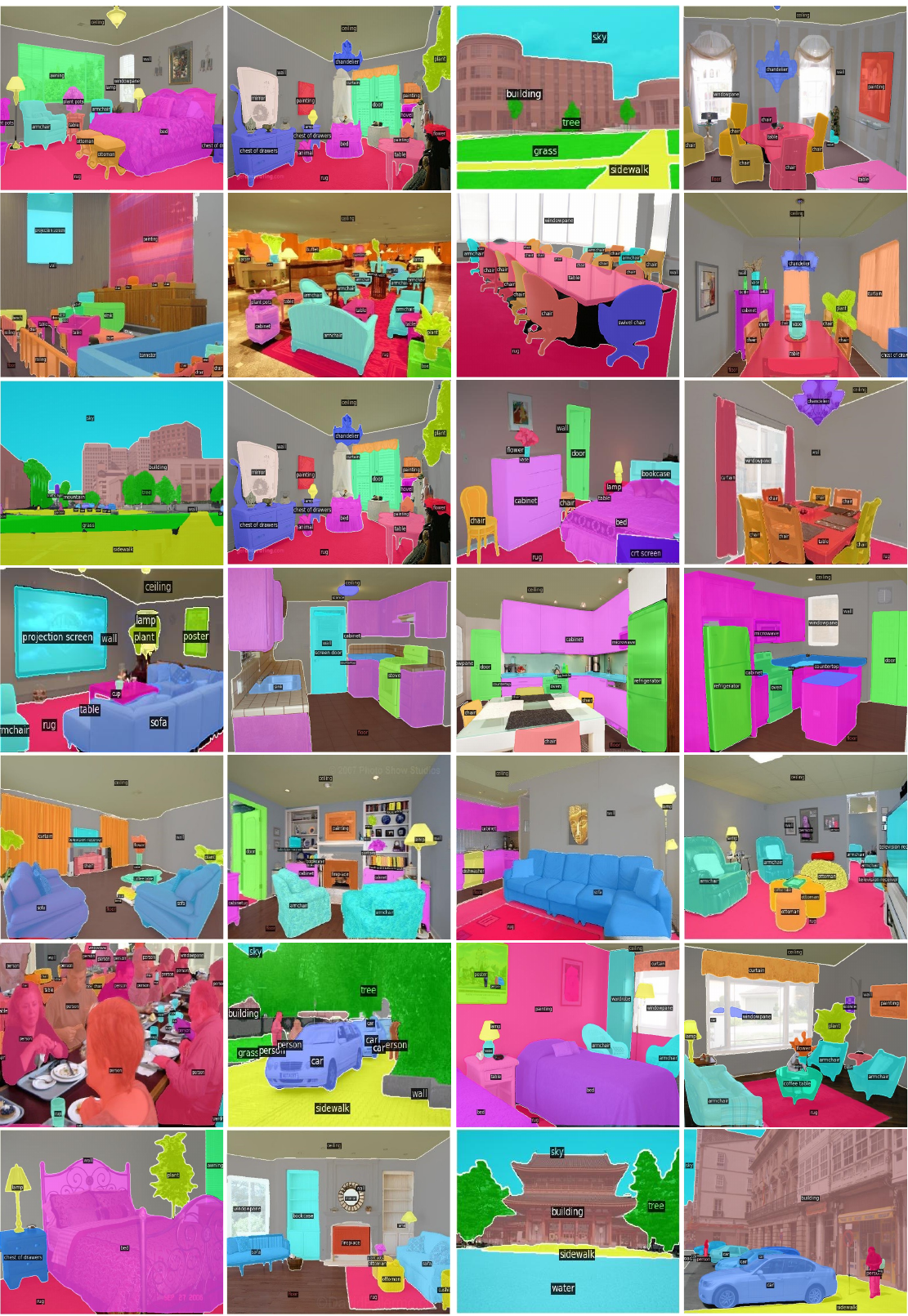}
    \vspace{- 1.0 em}
    \caption{PosSAM visualization of open-vocabulary panoptic segmentation predictions on ADE20K}
    \label{fig:suppli_ade1}
\end{figure}

\begin{figure}
    \centering
    \includegraphics[width=0.88\linewidth]{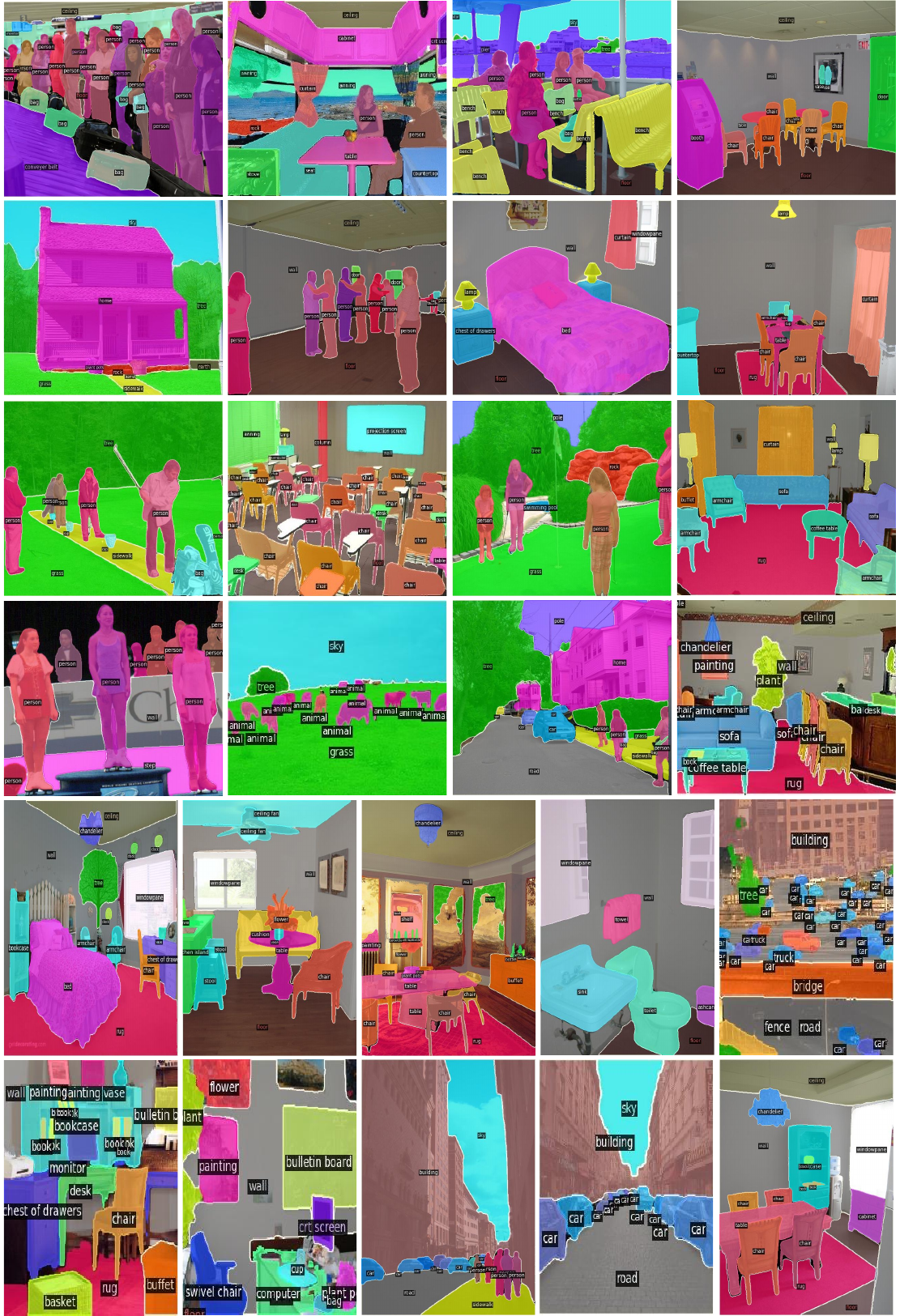}
    \vspace{- 0.05 em}
    \caption{PosSAM visualization of open-vocabulary panoptic segmentation predictions on ADE20K }
    \label{fig:suppli_ade2}
\end{figure}

\section{Open-Vocabulary Visualization}
\label{sec:suppli_vis}
To demonstrate the open-vocabulary recognition capabilities of PosSAM, we present extensive visualization for open-vocabulary panoptic segmentation predictions on ADE20K in Figure \ref{fig:suppli_ade1} and Figure \ref{fig:suppli_ade2}.

\section{Limitations and Future Work}
\label{sec:suppli_limit}
In our current work, we aim to enhance instance-aware masks by leveraging SAM and the proposed LDP and MASE modules to robustly perform real-world open-vocabulary segmentation. However, like prior OV methods \cite{yu2023convolutions,li2024omg,xu2023odise,zou2023generalized,liang2023open}, we still rely on CLIP backbone.  Ideally, a single backbone that embodies both spatial and semantic awareness would be optimal. Such an integration could not only reduce the number of parameters but also potentially enhance performance significantly, presenting a promising direction for future research.
\end{document}